# Agentic AI-powered flexible fiber-bundle endoscopy for high-resolution NIR-II fluorescence imaging in vivo

Yanzhao Shi[1,2,†], Yuanhua Liu[1,2,†], Sixin Xu[1,†], Wayne Jason Li[3], Yuyuan Chen[1], Danyang Xu[1], Zhisheng Wu[1], Hanze Yu[1], Ian Yu-Hong Wong[4], Simon Ying-Kit Law[4], Hongjie Dai[2,3,5,*], Liangqiong Qu[6,*], and Feifei Wang[1,2,*]

[1] Department of Electrical and Computer Engineering, School of Biomedical Engineering, The University of Hong Kong, Hong Kong SAR, China.

[2] Materials Innovation Institute for Life Sciences and Energy (MILES), The University of Hong Kong Shenzhen Institute of Research and Innovation (HKU-SIRI), Shenzhen 518045, China.

[3] Department of Mechanical Engineering, The University of Hong Kong, Hong Kong SAR, China.

[4] Department of Surgery, School of Clinical Medicine, LKS Faculty of Medicine, The University of Hong Kong, Hong Kong SAR, China.

[5] JC STEM Lab of Nanoscience and Nanomedicine, Department of Chemistry and School of Biomedical Sciences, The University of Hong Kong, Hong Kong SAR, China.

[6] School of Computing and Data Science, The University of Hong Kong, Hong Kong SAR, China.

† These authors contribute equally to this work.

* Correspondence to: hjdai@hku.hk, liangqqu@hku.hk, feifwang@hku.hk

**Abstract:**

Fiber-bundle endoscopy offers a compact and flexible route for clinical fluorescence imaging through natural human orifices, but since its first report in the 1950s, it has remained limited by low spatial resolution, honeycomb artifacts, and inter-core crosstalk. The crosstalk becomes more pronounced at near-infrared-II wavelengths (NIR-II, 1000-3000 nm), a spectral window that offers superior contrast, resolution, and tissue penetration depth for biomedical imaging. Here, we present an AI-powered flexible endoscopy platform that overcomes these constraints through optical-computational co-design: optimizing ultrathin fiber bundles to mitigate crosstalk-induced image blur and enable high-fidelity image transmission across the visible-to-NIR-II spectral range, and developing an Agent-Guided Mixture-of-Experts (GAME) pipeline for honeycomb-artifact removal and image restoration. GAME provides a single restoration entry point for diverse biomedical images acquired with our endoscope, spanning cell, mouse and human samples. It dynamically routes each input to suitable restoration experts via a vision-language model, facilitating image reconstruction with a fourfold resolution improvement beyond the Nyquist-Shannon sampling limit. The utility of our endoscope is demonstrated through in vivo NIR-II imaging of anatomical structures in mice, as well as imaging of the digital micromirror device (DMD)-projected human gastric tube and lymphatic system, paving the way for future clinical translation.

## Introduction

Flexible endoscopy is recommended as the first-line diagnostic and therapeutic option for a wide variety of clinical scenarios [1, 2]. Its compact design enables non-invasive navigation through narrow, tortuous lumens, including the esophagus (~20 mm) [3, 4], small intestine (~25-30 mm) [5, 6] and ureter (~3-4 mm) [7, 8], while providing real-time visualization for disease assessment. Currently, clinical flexible electronic endoscopes, such as gastrointestinal endoscopes, use miniature lenses and cameras integrated into the distal tip to achieve high-resolution morphological white-light imaging (Supplementary Table 1). However, these standard systems typically lack the optical hardware necessary for fluorescence imaging owing to distal-tip size constraints [9].

Compared with flexible endoscopes, clinical rigid endoscopes, such as laparoscopes, can perform near-infrared-I (NIR-I, 700-900 nm) fluorescence imaging using indocyanine green (ICG; peak emission: ~815 nm), a contrast agent approved by the U.S. Food and Drug Administration (FDA), with fluorescence signals relayed from the distal tip to a proximal camera through a complex lens system. Such rigid endoscopes are now routinely used in minimally invasive urologic[10], gynecological [11], hepatobiliary [12], and pancreatic [13] surgeries to enhance intraoperative decision-making by visualizing fluorescently labeled anatomical structures and pathological lesions [14, 15]. Recently, near-infrared-II (NIR-II, 1000-3000 nm) fluorescence imaging [16] has emerged as a promising alternative for clinical applications [17-20], offering deeper tissue penetration, higher contrast, and improved spatial resolution than conventional NIR-I imaging due to reduced light scattering and diminished autofluorescence at longer wavelengths. Accordingly, NIR-II rigid endoscopes [21, 22] have shown superior imaging performance over NIR-I approaches in preclinical animal studies.

These advances highlight the potential value of extending fluorescence imaging, particularly NIR-II fluorescence imaging, into flexible endoscopic platforms to further augment their clinical functionality [9]. Flexible fiber-bundle endoscopy (FBE) offers an attractive approach for fluorescence imaging as it can provide a larger field-of-view (FOV) and more intuitive endoscopic navigation than other fiber-optic methods (Supplementary Table 2). However, despite more than 70 years of development since its first introduction in the 1950s [23], FBE still faces fundamental limitations [24] that have constrained its broader use. First, honeycomb artifacts and low resolution are induced by downsampling through a finite number of discrete fiber cores, typically ~$10^3$-$10^5$. Second, inter-core crosstalk caused by optical leakage and coupling between adjacent fibers degrades fine image features [25, 26]. Third, the limited number of cores makes the trade-off between FOV and resolution more severe, such that under clinically preferred FOV conditions (e.g., FOV = 10 cm; core number = 10,000), the effective resolution degrades to the millimeter scale. Although FBE has recently been explored for ICG- and bevacizumab-800CW-based NIR-I imaging [27, 28], its resolution remains markedly lower than that of electronic endoscopes. Extending FBE fluorescence imaging into the NIR-II window further aggravates this challenge, because crosstalk becomes more pronounced at longer wavelengths, making image transmission impossible [26].

Currently, AI networks have been explored for visible-light fiber-bundle image restoration, mainly relying on synthetic paired data [29], physical prior knowledge [30], or unsupervised learning strategies [31]. However, these approaches remain insufficient for capturing real degradation physics and reconstructing heterogeneous biomedical images, which vary substantially in scale and morphology, from subcellular structures to mouse tissues and human organs. For such diverse biomedical scenarios, a general model trained across all these image types can offer broad coverage [29, 32] but may perform unevenly because optimal restoration strategies vary with image content. Conversely, specialized models provide stronger domain-specific restoration but require prior knowledge of imaging conditions and scale poorly to new biomedical scenarios. These limitations constrain high-resolution reconstruction and the application of fiber-bundle imaging in real-world settings, where the tissue type, fluorescence window, and degradation pattern may be unknown before imaging.

Here, we present an AI-powered flexible endoscopy platform based on optical-computational co-design, integrating a broad-spectrum fiber-bundle imaging system spanning the visible to NIR-II range with an agentic AI reconstruction network. Optically, we optimized the fiber configuration to mitigate image degradation induced by inter-core crosstalk, achieving high-fidelity image transmission across the 485-1550 nm spectral window. Computationally, we developed GAME, an Agent-Guided Mixture-of-Experts reconstruction framework that makes fiber-bundle image restoration adaptive, confidence-aware and extensible. GAME first generates a preliminary restoration to suppress honeycomb artifacts, then uses a vision-language agent to analyze the restored image and route the input to appropriate detail-refinement experts, and finally selects the most reliable reconstruction through reliability assessment. We demonstrated that GAME removes intrinsic honeycomb artifacts and restores fine details beyond the native core-size limit across diverse biomedical images recorded by our endoscope, spanning cells, mouse and human tissues, subcellular-to-organ scales and visible-to-NIR-II wavelengths. GAME is designed for extensibility: ambiguous or unseen cases can be flagged through confidence-aware routing and reliability assessment, and new experts can be incorporated without retraining the existing pipeline, enabling the system to evolve as new biomedical and clinical applications emerge. We demonstrated the utility of GAME-powered endoscopy through in vivo imaging of the mouse intestine, liver, ureter, and lymphatic system, as well as imaging of digital micromirror device (DMD)-projected patterns of the human gastric tube and lymphatic system.

## Results

### Agentic AI-powered flexible fiber-bundle endoscopy platform

Our home-built AI-powered FBE system was designed as an optical-computational platform that comprises a fiber-bundle endoscope for fluorescence imaging from the visible to NIR-II windows (Fig. 1a), and an AI-powered, agent-guided mixture-of-experts pipeline for fiber-bundle image restoration (Fig. 1b). The optical module enables fluorescence imaging across the visible, NIR-I and NIR-II windows, whereas the computational module restores fiber-transmitted images by suppressing honeycomb artifacts and recovering fine biological structures beyond the native sampling limit of the fiber bundle.

The FBE was assembled by integrating a miniature lens with a field of view of 60° at the distal tip of an ultrathin fiber bundle with a diameter of 0.6, 1.4, or 3.2 mm (Supplementary Fig. 1). These diameters are smaller than those of many commercially available clinical endoscopes, such as rhinoscopes, falloposcopes, and cystoscopes (Fig. 1a and Supplementary Table 1), thereby facilitating fluorescence imaging in confined natural human lumens. The proximal end of the fiber bundle was coupled to a wide-field microscope equipped with a 5× or 10× objective, selected according to the fiber bundle diameter, and a 200-mm tube lens. Fluorescence signals were collected using a silicon-based camera and an indium gallium arsenide (InGaAs) camera, enabling fluorescence imaging across the visible, NIR-I, and NIR-II spectral windows (Methods). The excitation light can be delivered through either an extra illumination fiber (Fig. 1a) or the imaging fiber, depending on different applications. The raw images recorded by the fiber-bundle endoscope contain honeycomb artifacts that degrade image quality and limit effective resolution.

Restoring raw fiber-bundle images into high-fidelity images across diverse biomedical scenarios is challenging because uncertainties arising from biological structural randomness, imaging scale variations, fiber-bundle artifacts, and signal-dependent noise interact differently with various biological morphologies and imaging conditions. We tackled these issues by designing an adaptive agent-guided restoration framework (Fig. 1b and Supplementary Fig. 2) and constructing a broadly representative paired training dataset, thereby facilitating effective image reconstruction (Fig. 1c).

At the algorithmic level, we developed GAME, a self-evolving agent-guided mixture-of-experts framework that automatically analyzes each input and routes it to the suitable restoration experts (Fig. 1b, Methods). GAME operates in three steps: (1) a general restoration model first suppresses honeycomb artifacts and recovers coarse biological structure; (2) a vision-language model (VLM) then analyzes this intermediate result to determine the image category and routes it to the suitable specialized experts; and (3) a reliability module evaluates candidate outputs, selects the optimal reconstruction, and reports a confidence score that provides an explicit measure of reconstruction reliability. This design addresses the trade-off between broadly trained general models, which are robust but may miss fine details, and narrowly specialized models, which can achieve high fidelity but generalize poorly. Crucially, the framework is extensible: when a new imaging scenario is encountered (e.g., a new tissue type), a new expert can be incorporated without retraining the existing system, enabling the platform to grow progressively as clinical applications expand (Supplementary Fig. 3).

At the data level, a key requirement is providing paired supervision that captures the diversity of biomedical fiber-bundle imaging. Direct acquisition of matched clean and fiber-degraded images across different scenarios is impractical, as clean counterparts are often unavailable in real endoscopic measurements, and fiber-induced degradation depends on both image content and optical configuration. We therefore established a DMD-based image acquisition workflow (Fig.

2a,b and Methods) to generate matched pairs at different wavelengths using LEDs with distinct emission wavelengths, mimicking fluorescence imaging from the visible to the NIR-II window. The generated image can be directly captured by the miniature lens-mounted fiber-bundle endoscope, or collected by a tube lens and then demagnified onto the distal end of the fiber bundle by an objective, with the transmitted image from the proximal end collected by a wide-field microscope (Fig. 2a,b and Methods). In each case, the clean image input to the DMD served as the ground truth, whereas the fiber-transmitted image provided the corresponding degraded counterpart with realistic honeycomb artifacts. Using this system, we acquired 96,050 paired images covering cells, mouse tissues and human tissues, with feature scales from 0.1 μm to 10 cm and wavelength from 485-1550 nm across the visible, NIR-I, and NIR-II windows (Supplementary Table 3).

We validated GAME using a representative NIR-IIb (1500-1700 nm) fluorescence image of microvascular networks in the mouse head (Fig. 1c). This NIR-IIb image was recorded ~5 min post intravenous injection of core/shell lead sulfide/cadmium sulfide (PbS/CdS) quantum dots. Emission was filtered using a 1500-nm long-pass filter, and an 808-nm laser was used for excitation. The raw fiber-bundle image was severely obscured by honeycomb patterns, making the blood vessels difficult to recognize. Gaussian filtering caused blurring; U-Net [33] failed to resolve fine capillary connectivity; and the recent UniFMIR [32] model improved contrast but retained structural distortions and low resolution. In contrast, GAME effectively eliminated artifacts and restored fine vascular details closely matching the ground truth, achieving the highest PSNR and SSIM, the lowest LPIPS, and the best resolution among all methods (Fig. 1c).

**Mitigation of inter-core crosstalk across the visible to NIR-II windows**

Inter-core crosstalk, arising from evanescent field coupling between adjacent cores [25], degrades spatial resolution and blurs images in fiber-bundle-based imaging. The strength of this crosstalk is influenced by the operating wavelength ($\lambda$) [26], numerical aperture (NA) [34], core diameter ($D_{\mathrm{core}}$), and core-to-core pitch (*Pitch*) [34]. We first compared the image-transmission performance of three fibers with different configurations (Fiber 1, Fiber 2, and Fiber 3; Fig. 2a) at imaging wavelengths spanning the visible, NIR-I and NIR-II windows using the DMD-based imaging system (Fig. 2b, Methods).

For Fiber 1, with a $D_{\mathrm{core}}$ of 2.5 μm, a *Pitch* of 3.1 μm and an NA of 0.39, the raw transmitted image became increasingly blurred because of crosstalk as the imaging wavelength increased from 485 nm to 850 nm and into the NIR-II window (Fig. 2c and Supplementary Fig. 4). No detailed structures were observed when $\lambda \geq 850$ nm, and only speckle patterns were recorded when $\lambda \geq 1350$ nm. As the wavelength increases, the normalized frequency, $V = \pi \mathrm{NA} D_{\mathrm{core}}/\lambda$, and the total number of modes, $N \approx V^2/2 = (\pi \mathrm{NA} D_{\mathrm{core}}/\lambda)^2/2$, of Fiber 1 decrease [35], with $V$ and $N$ decreasing from 6.32 and ~20 at 485 nm to 1.98 and ~2 at 1550 nm, respectively (Supplementary Fig. 1). Despite this reduction in $V$ and $N$, image transmission deteriorates markedly at longer wavelengths because the modal field becomes less confined within individual cores and extends further into the cladding/surrounding medium [36], which increases inter-core crosstalk and compromises spatial fidelity.

We then reconstructed raw images acquired through Fiber 1 at different wavelengths, which contained honeycomb artifacts or speckle patterns, using different processing methods, including Gaussian filtering, U-Net, UniFMIR and our method (Fig. 2c,d and Supplementary Figs. 4 and 5). At 485 nm, our network effectively suppressed honeycomb artifacts, achieving the highest fidelity to the reference images. At 850 nm, it recovered structural features that were barely discernible in the raw images and were poorly reconstructed by other methods. In the NIR-II window, only coarse structures or the mouse outline could be reconstructed by AI-based methods, whereas no effective structural information was recovered from the traditional Gaussian-filtered results. Quantitative results in Supplementary Fig. 5 further showed that our method achieved higher PSNR and SSIM, and lower LPIPS, than Gaussian filtering, U-Net and UniFMIR from 485 to 1050 nm, revealing improved robustness at longer wavelengths. Beyond 1050 nm, however, severe crosstalk-induced degradation challenged all models, narrowing the performance gap among the different methods. These results indicate that crosstalk-induced blur strongly limits image reconstruction effectiveness and quality.

To suppress inter-core crosstalk and achieve efficient image transmission in the NIR-I and NIR-II windows, we increased the core diameter and NA to 8 μm and 0.49, respectively, for Fiber 2 and to 12 μm and 0.55, respectively, for Fiber 3, while also increasing the separation between neighboring cores (Fig. 2a and Supplementary Fig. 1). These changes raise $V$, enhance modal confinement within each core, and reduce evanescent overlap with neighboring cores. However, they also increase the number of supported modes and may affect the spatial sampling density if the fiber bundle outer diameter remains unchanged. To minimize the influence of core number on imaging resolution, we used Fiber 2, which had a core number similar to that of Fiber 1, to compare image transmission performance at different wavelengths (Fig. 2c and Supplementary Figs. 4-7). Compared with Fiber 1, Fiber 2 enabled more effective image transmission from the visible to the NIR-II window by reducing inter-core-crosstalk-induced blur, facilitating the observation of detailed structures in the transmitted image. After reconstruction of these raw images containing honeycomb artifacts using different processing methods, our GAME model achieved higher-resolution reconstruction of detailed structures than the other algorithms (Fig. 2c and Supplementary Fig. 6), yielding the highest PSNR and SSIM values and the lowest LPIPS values across the 485-1550-nm wavelength range (Supplementary Fig. 7).

Fiber 3 has a larger core diameter, higher NA, and greater number of cores than Fiber 2 (Fig. 2a and Supplementary Fig. 1). With all fiber cores used for imaging, Fiber 3 delivered more precise structural details and preserved structural integrity in the raw image, thereby facilitating the reconstruction of high-quality images by our computational algorithms at operating wavelengths in the visible, NIR-I, and NIR-II windows (Fig. 2c and Supplementary Figs. 8 and 9). Quantitative metrics, including PSNR, SSIM, and LPIPS (Fig. 2e), corroborate these visual observations: Fiber 1 exhibits sharp performance declines at longer wavelengths, whereas Fibers 2 and 3 remain robust across the studied spectrum.

## Sub-core-resolution imaging beyond the theoretical sampling limit

Assuming that the fiber bundle is used directly for imaging, its imaging resolution is limited to ~2*Pitch*, according to the Nyquist-Shannon sampling theorem. The imaging performance of a fiber bundle can be evaluated by the feature size of the transmitted image normalized by the core diameter ($S_F/D_{core}$) or pitch ($S_F/Pitch$), where $S_F$ denotes the feature size. These dimensionless ratios eliminate the influence of FOV variations arising from the use of front lenses with different magnifications and reflect the effective imaging resolution relative to the intrinsic sampling scale of the fiber bundle. To evaluate the resolution limit of our restoration model, we first tested its ability to recover mouse blood vessels with different $S_F/D_{core}$ and $S_F/Pitch$ values (Fig. 3a). The dataset was generated using the DMD imaging system equipped with Fiber 2 (Fig. 2a) and a 1550-nm LED (Fig. 2b). The $S_F/D_{core}$ and $S_F/Pitch$ values of the blood vessels were varied by scaling the ground-truth image input to the DMD with magnification factors of 1.0×, 1.4×, and 2.0×. As this scaling factor increased, $S_F/D_{core}$ and $S_F/Pitch$ values of the blood vessels also increased, meaning that more fiber cores sampled the same vascular structure (Fig. 3a).

Different restoration methods were applied to reconstruct images from raw data containing honeycomb artifacts (Fig. 3a). Across all magnifications, our network yielded images with the highest fidelity to the ground-truth images, as evidenced by visual observation (Fig. 3a), higher PSNR and SSIM values and lower LPIPS values (Supplementary Fig. 10). In addition, as the scaling factor increased, the reconstruction performance of all methods improved, because more fiber-core information was available for restoring the same feature. Quantitative results showed that the $S_F/D_{core}$ and $S_F/Pitch$ values of the smallest restorable blood vessels remained unchanged as the magnification increased from 1.0× to 2.0×, indicating that the minimum resolvable feature of the GAME-enhanced fiber-bundle system was intrinsically limited to ~ 0.75 times the core diameter and ~ 0.59 times the pitch for Fiber 2 at 1550 nm (Fig. 3b), thereby demonstrating sub-core-resolution imaging and a ~3.4-fold improvement over the theoretical sampling limit. Applying higher magnifications shifts smaller anatomical details beyond this fixed decoding threshold, thereby enabling more precise reconstruction and enhancing the absolute resolution.

We further evaluated the $S_F/D_{core}$ and $S_F/Pitch$ values of the smallest restorable blood vessels across different fiber configurations and imaging wavelengths (Fig. 3c). These results demonstrate our model's ability to recover high-frequency spatial information, with $S_F/Pitch$ values remaining ~0.5-0.75 for all tested fibers, wavelengths, and magnifications (Fig. 3b,c), representing a ~ 2.7- to 4-fold improvement over the resolution limit imposed by the Nyquist-Shannon sampling theorem. Therefore, the information coupled into fiber cores is sufficient for our algorithm to reconstruct blood vessels with diameters smaller than that of an individual core. We hypothesize that the AI network achieves this by exploiting sub-core intensity variations [37] and modal crosstalk patterns [38, 39], thereby decoding the rich spatial and angular information inherently preserved in the image transmitted through the fiber bundle.

**In vivo AI-powered flexible fiber-bundle endoscopy in the NIR-I and NIR-II sub-windows**

To evaluate the practical applicability of flexible FBE and the generalization capability of our GAME network in real-world scenarios, we transitioned from DMD-projected images to in vivo NIR-I and NIR-II fluorescence imaging of mice administered with ICG (Fig. 4a,b). We installed a custom miniature lens with a field of view of 60° at the distal tip of a fiber-bundle endoscope fabricated using Fiber 2 (Fig. 2a). The proximal end of fiber bundle was coupled to a wide-field microscope equipped with a 10× objective for NIR-I and NIR-II fluorescence imaging (Methods). An 808-nm laser was used for ICG excitation. A key challenge in real-world intraoperative settings is the lack of strictly paired, degradation-free ground-truth data, precluding standard quantitative evaluation with metrics such as PSNR, SSIM and LPIPS. To assess the model output, we employed a wide-field imaging system to acquire reference images of the same anatomical regions observed by FBE, which served as ground-truth images.

We first performed NIR-II imaging of the mouse intestine and liver at 2 hours (Fig. 4c) and 3-4 hours (Fig. 4d) after retro-orbital intravenous administration of ICG, respectively. Fluorescence emission was collected after passing through a 1100-nm long-pass filter for intestine imaging and a 1000-nm long-pass filter for liver imaging. Honeycomb artifacts were observed in the raw endoscopic data, interfering with the inspection of anatomical structures in the intestine and liver. We then processed these raw images using our restoration model. We note that our network had not been specifically trained on these intestine and liver datasets. The model effectively eliminated the honeycomb artifacts present in the raw inputs and faithfully reconstructed organ outlines, such as intestinal loops and liver boundaries, closely matching the wide-field ground-truth images (Fig. 4c,d).

To validate the reconstruction of anatomical structures with feature sizes close to the core diameter of the fiber bundle, we conducted NIR-I fluorescence imaging of the mouse ureter (Fig. 4e). For ureter labeling, we retrogradely injected ICG through the urethra, referencing a procedure used for ICG-guided ureter localization in human surgery [40]. NIR-I fluorescence was collected through an 830-nm long-pass filter. Notably, despite the absence of training data on these specific ureteral morphological features, our model effectively eliminated severe honeycomb artifacts and reconstructed the fine ureters connecting the kidney and bladder from low-contrast raw images (Fig. 4e). Furthermore, line profile analysis of the restored ureter yielded a full-width at half-maximum (FWHM) of 0.30 mm, comparable to the 0.29 mm measured from the wide-field ground truth.

Fluorescence-guided visualization of lymph nodes and lymphatic vessels during tumor surgery enables identification of drainage pathways, sentinel lymph nodes, and potential metastases, thereby guiding precise resection while minimizing tissue damage [16, 41]. Motivated by this clinical relevance, we further evaluated our AI-powered flexible FBE for NIR-II imaging of mouse lymph nodes and lymphatic vessels 30 min after subcutaneous injection of ICG at the base of the tail (Fig. 4f). The fluorescence signal was collected through a 1000-nm long-pass filter. Consistent with our previous observations, the restoration pipeline recovered the contours of the lymph nodes and enhanced the visibility of the lymphatic vessels. Line profile analysis yielded an FWHM of 0.38 mm for the restored lymphatic vessel closely matching the wide-field reference (0.35 mm). These results indicate the model's capacity to resolve delicate tubular structures in anatomical environments that were not included in the training dataset.

In clinical practice, overlaying pseudo-colored fluorescence images onto white-light images can help localize contrast-agent-labeled targets within tissue. To demonstrate the translational potential of our flexible FBE for surgical guidance, we merged the computationally restored NIR-I and NIR-II fluorescence images with corresponding white-light images (Fig. 4c-f, row 4). Co-registering the restored fluorescence images with color brightfield images provides complementary morphological context, anchoring the functional NIR-I and NIR-II signal to visible anatomical landmarks. These results confirm that our approach exhibits robust generalization in complex, GT-free environments, highlighting its significant potential for clinical diagnostics and intraoperative navigation.

We further compared our method with Gaussian filtering, U-Net and UniFMIR on the same unseen in vivo FBE data (Supplementary Fig. 11). Gaussian filtering removed the honeycomb pattern but blurred anatomical details, whereas U-Net left residual artifacts and weakened fine structural signals. UniFMIR showed limited effectiveness on data, possibly because its Swin Transformer-based feature enhancement had insufficient generalization to these out-of-distribution in vivo images. In contrast, our method consistently removed honeycomb artifacts while preserving organ boundaries and fine tubular structures, showed the closest visual agreement with the wide-field references, and demonstrated robust adaptation to previously unseen in vivo images.

**GAME-based robust cross-scale image restoration via autonomous expert routing**

Fiber bundles serve as image transmission media and have been explored for imaging cells, mouse tissues, and human organs [24, 27, 28, 34, 37, 42]. In Figures 2 and 3, we demonstrated that our GAME network can robustly restore sub-core-scale features of mouse blood vessels. To further validate the generalizability and practical potential of the GAME pipeline, we evaluated its predictive performance across a broad range of biological scales, from subcellular organelles (~100 nm) to macroscopic human organs (~10 cm) (Fig. 5a). We directly fed raw images from diverse categories into GAME, allowing the pipeline to autonomously infer the image type, route the input to appropriate experts, and generate the final restoration. For each imaging scenario, we systematically visualized both the final restored images and the intermediate decision-making processes of the agent, including VLM-based anatomical recognition, dynamic expert-model allocation, and terminal reliability evaluation (Fig. 5b-h, Supplementary Fig. 12).

Specifically, when a raw FBE image, such as an image of a cell nucleus (Fig. 5c), was input into the GAME pipeline, a general structural restoration model first generated a preliminary restored image by removing honeycomb artifacts (Supplementary Fig. 13). The VLM then analyzed this preliminary image, predicted its category, and assigned category-specific confidence scores, for example 0.82 for cell, 0.15 for mouse tissue, 0.03 for human tissue, and 0.00 for an unknown category that included out-of-domain samples unidentifiable by the VLM. Using a predefined confidence threshold of 0.75 to guide restoration-expert allocation, the pipeline determined that the confidence score for the cell category was the highest and exceeded the threshold; and therefore, it directly allocated a single expert model, the Cell model, to restore the detailed structures. As a result, the nucleus image was reconstructed with high fidelity. We evaluated the reliability score of this reconstructed image by extracting and comparing its global structural information with that of the preliminary predicted image (Supplementary Fig. 2). The resulting score of 0.93 demonstrates high structural fidelity in the final restoration.

For ambiguous targets such as mouse brain vasculature (Fig. 5e, Supplementary Fig. 14), which visually resembles human vasculature, the VLM assigned a confidence score of 0.72 to mouse, 0.18 to human, 0.15 to cell, and 0.05 to unknown category. Although mouse category had the highest confidence score, it was below the predefined threshold of 0.75. This uncertainty triggered parallel allocation to the experts corresponding to the two highest-confidence categories, Mouse and Human for this case, to better ensure inclusion of the correct expert. The pipeline then computed a reliability score for each expert reconstruction. The Mouse expert yielded a higher reliability score than the Human expert (0.88 vs. 0.81), indicating fewer structural hallucinations, and was therefore autonomously selected as the final output. This self-correcting function enables the pipeline to robustly handle unpredictable inputs.

We further evaluated whether this autonomous routing strategy could yield the high-fidelity image restorations required for clinical translation. Two human intraoperative datasets were used to challenge the GAME pipeline. First, we assessed the restoration of NIR-I images of human gastric conduits during esophagectomy after ICG administration (Fig. 5g). Clear visualization of gastric conduit perfusion during esophagectomy is essential for optimizing the anastomotic site and preventing life-threatening anastomotic leakage [17]. GAME successfully reconstructed perfusion patterns in the gastric conduit, revealing continuous spatial perfusion features that were largely obscured in the raw fiber-bundle images. We note that GAME reconstructed blood vessels that were missed by U-Net and UniFMIR, as indicated by the arrows in Supplementary Fig. 12, demonstrating high-fidelity recovery and preventing the loss of important information.

The second case involved NIR-II imaging of human lymph nodes and lymphatic vessels during surgery for adenocarcinoma of the stomach after subcutaneous injection of ICG (Fig. 5h). Fluorescence was collected through a 1050 ± 25 nm bandpass filter. After input into the GAME pipeline, this image was assigned to the Human expert model for reconstruction, yielding a reliability score of 0.98 for the final result. The reconstructed image in Fig. 5h shows effective elimination of honeycomb artifacts, enabling clear demarcation of human lymph nodes and lymphatic vessels. As indicated by the arrows in Supplementary Fig. 12, GAME reconstructed a lymphatic vessel connected to the lymph node, consistent with the ground-truth image, whereas U-Net and UniFMIR failed to reconstruct this structure. Such reconstruction failure may result in missed resection of sentinel lymphatic vessels, potentially increasing the risk of cancer recurrence. When we overlay the fluorescence image onto the white-light image, we can locate the lymph nodes and lymphatic vessels within the tissue environment. These clinical examples show the potential of our AI-powered flexible FBE for use in clinical scenarios.

Further quantitative results across 20 diverse test sets spanning cell, mouse-tissue, and human-tissue images confirmed the advantage of GAME over Gaussian filtering, U-Net, and UniFMIR, achieving higher PSNR and SSIM and lower LPIPS (Supplementary Fig. 15). In particular, GAME reduced LPIPS on cell datasets by ~ 0.10 compared with U-Net and UniFMIR, and improved SSIM on human-tissue datasets by ~ 0.10, demonstrating robust restoration of both fine subcellular structures and clinically relevant anatomical features. These results further indicate that GAME generalizes effectively across diverse biomedical imaging scenarios.

## Conclusion

Resolution degradation caused by the intrinsic downsampling of fiber bundles has remained a persistent challenge since their introduction into endoscopy in 1954 [23] and was a major factor in their replacement by electronic endoscopes [43]. Although methods such as fiber-bundle shifting[42] and spectral coding [44] have been proposed to overcome this limitation, they increase system complexity, can introduce motion artifacts, enlarge device size, and may not be compatible with fluorescence imaging. To address this challenge, we developed the GAME framework (Fig. 1b and Supplementary Fig. 2), trained and evaluated with 96,050 acquired image pairs, which enables computational sub-Nyquist spatial recovery of features as small as ~0.75 times the fiber-core diameter, yielding up to a four-fold improvement in effective resolution without hardware alterations (Fig. 3).

Through a self-evolving agent pipeline that decouples global structural recovery from adaptive expert-driven detail restoration, GAME reconstructed diverse biomedical targets, ranging from cells to mouse and human tissues, with higher fidelity than classical spatial-filtering and other advanced deep-learning methods. For example, GAME reconstructed blood vessels in a human gastric conduit and sentinel lymphatic vessels during adenocarcinoma surgery that were missed by U-Net and UniFMIR (Supplementary Fig. 12), thereby facilitating more precise surgical guidance for potential biomedical applications. GAME can effectively overcome the trade-off between broadly trained general models and narrowly specialized restoration networks [29, 30, 31, 32], enabling more robust adaptation to previously unseen in vivo images than UniFMIR (Supplementary Fig. 11). Reliability evaluation was further incorporated to reduce the risk of generative hallucinations and improve reconstruction robustness. Beyond performance, GAME provides an adaptive, extensible, and user-steerable expert-orchestration framework for practical deployment: newly trained experts can be seamlessly integrated to accommodate additional image categories without rebuilding the entire system, while restoration can be flexibly driven by automatic routing, user free-text guidance, or direct expert invocation.

To enable flexible endoscopic fluorescence imaging, particularly in the NIR-II window, we optimized the fiber-bundle design by increasing the core diameter, NA and core-to-core spacing relative to conventional bundles designed for visible-light imaging. These modifications improve modal confinement within individual cores and suppress wavelength-dependent inter-core crosstalk, which becomes increasingly severe at longer wavelengths in conventional small-core, low-NA fiber bundles. As a result, the optimized fiber bundles achieved high-fidelity image transmission across the visible, NIR-I and NIR-II spectral windows (Fig. 2 and Supplementary Figs. 4-9). The image-relay properties of a fiber bundle can greatly relax the size constraints on NIR-II cameras in a compact imager, such as an endoscope, as these cameras are typically large because they require a cooling system to reduce thermal noise. When empowered by the GAME network, this design enables high-resolution, flexible fluorescence endoscopy over a broad spectral range. We demonstrated the utility of our NIR-II endoscopic platform through in vivo imaging of the mouse intestine, liver, ureter, and lymphatic system, as well as imaging of DMD-projected cellular structures, the human gastric tube and the lymphatic system.

Therefore, our AI-powered flexible fiber-bundle endoscope mitigates the long-standing limitations of conventional FBEs, including honeycomb artifacts, limited spatial resolution, inter-core crosstalk, and the trade-off between FOV and resolution, thereby expanding its potential for high-resolution fluorescence imaging. Compared with rigid endoscopes constrained in navigating narrow, tortuous lumens and single-fiber systems limited by a small FOV (Supplementary Table 2), our FBE combines mechanical flexibility with high-resolution visible-to-NIR-II fluorescence imaging over a large FOV. Reconstructions of human lymph nodes, lymphatic vessels, and gastric conduits demonstrate its high resolution over clinically relevant FOVs (Fig. 5g,h). With an outer diameter down to 0.6 mm, the endoscope is smaller than many commercially available clinical endoscopes (Fig. 1a and Supplementary Table 1) and may facilitate minimally invasive fluorescence imaging of natural human lumens, such as the ureter and fallopian tube. In addition, it can be inserted through the tool channels of commercial gastroscopes or colonoscopes, enabling rapid fluorescence assessment during standard endoscopic procedures. For example, following gastric conduit reconstruction during esophagectomy, evaluation of the anastomotic site is critical because anastomotic leakage can be life-threatening[17]. In this scenario, our fiber-bundle endoscope can be introduced through the working channel of a commercial gastroscope to examine the anastomotic site and assess potential leakage. Our successful reconstruction of human gastric conduits paves the way for this potential clinical application. Extending fluorescence imaging to the NIR-II window can further improve tissue penetration depth, contrast, and spatial resolution in clinical applications [16].

In conclusion, we developed an agentic AI-powered flexible fiber-bundle endoscopy platform for high-resolution fluorescence imaging across the visible-to-NIR-II spectral windows. By integrating optimized fiber-bundle optics with reliable, expert-orchestrated AI reconstruction, this platform addresses key physical constraints that have historically limited fiber-bundle endoscopy and advances ultrathin NIR-II flexible fluorescence endoscopy toward clinical translation. Our approach has the potential to become a useful tool for fluorescence-guided diagnosis, intraoperative navigation and minimally invasive assessment of anatomical structures and pathological lesions.

## Data availability

All data supporting the findings of this study are presented in the main text and the Supplementary Information.

## Code availability

Our source code will be available upon acceptance.

## Methods

### Optical setup for paired data acquisition

We developed a DMD-based image projection platform (Fig. 2b) using a DMD module (V-650L NIR, ViALUX GmbH) to produce images at different wavelengths across the visible-to-NIR-II spectral range. LED sources with peak emission wavelengths of 485, 850, 1050, 1350, and 1550 nm illuminated the DMD to dynamically project high-resolution biological images, including cells, mouse tissues, and human tissues, thereby mimicking fluorescence imaging at different wavelengths. Three fibers with different configurations were used: Fiber 1, Fiber 2, and Fiber 3 (Fig. 2a and Supplementary Fig. 1). For each fiber, the projected image was collected by a 200-mm lens (AC254-200-C) and then demagnified onto the distal end of the fiber bundle by an objective with a magnification of 20× for Fiber 1, 10× for Fiber 2 and 4× for Fiber 3. After transmission through the corresponding fiber bundle, the image at the proximal end was collected by a wide-field microscope equipped with an objective with a magnification of 20× for Fiber 1, 10× for Fiber 2, and 5× for Fiber 3 and a 200-mm tube lens (TTL200-S8). The transmitted images were acquired using a silicon-based camera and an InGaAs camera. The images input to the DMD served as the uncorrupted ground-truth references.

### Dataset construction and acquisition

We constructed a paired dataset of clean ground-truth images and corresponding fiber-degraded images using the DMD-based image projection platform (Fig. 2b). The dataset included cellular images, mouse tissue images and human tissue images, spanning diverse biological structures and wavelengths from the visible to the NIR-II window (Supplementary Table 3). For each image pair, the clean image input into the DMD served as the ground truth, whereas the fiber-transmitted image provided the corresponding degraded input for model training and evaluation.

The cell subset was generated using established public datasets, including 3DRCAN [45], BioSR [46], GigaDB [47]. The mouse subset was generated using our previously reported datasets [48]. The human subset was constructed using previously published clinical data [17] and our own data from HKU Medicine.

To evaluate reconstruction resolution, mouse data were input to the DMD at six magnifications, ranging from 1.0× to 2.0× in 0.2× increments, to simulate diverse fields of view. Furthermore, to assess wavelength-dependent performance, the mouse data were acquired using three fiber bundles at 485, 850, 1050, 1350, and 1550 nm, covering the visible, NIR-I and NIR-II windows. In addition, to augment training diversity and account for the influence of the miniature lens on the network, we projected mouse images using a small monitor and recorded them with a fiber-bundle endoscope equipped with a miniature lens with a field of view of 60° at its distal end.

### GAME model architecture

High-fidelity restoration of raw fiber-bundle measurements across diverse biomedical scenarios is challenging because fiber-bundle transmission produces a non-stationary degradation process. Raw fiber-bundle images are corrupted by multiple interacting optical degradations, including honeycomb artifacts, inter-core crosstalk, and signal-dependent noise. The relative contributions of these degradations vary with biological morphology, imaging scale, and operating wavelength. This variability is particularly pronounced in NIR-II imaging, where longer wavelengths exacerbate inter-core crosstalk (Fig. 2c and Supplementary Figs. 4, 6, and 8).

For practical clinical deployment, a restoration model must therefore generalize across these heterogeneous conditions without prior knowledge of the specific imaging scenario. This requirement creates a fundamental tension in model design. A single restoration model [29, 32] trained on diverse biomedical scenarios can provide broad coverage, but may perform unevenly because the optimal restoration strategy differs across image contents, spatial scales and operating wavelengths; conversely, specialized models can achieve stronger domain-specific restoration, but require the imaging condition to be known in advance and are difficult to scale to new biomedical scenarios. This problem is further compounded by the scarcity of real paired training data. Existing AI-based fiber-bundle image restoration methods have used synthetic paired data [29], physical priors [30], or unsupervised learning strategies [31] to alleviate the data scarcity issues. However, these approaches still provide insufficient exposure to real degradation physics and cannot adequately model wavelength-dependent crosstalk across diverse tissue types.

To address these challenges, we developed GAME, a self-evolving agent framework that reframes fiber-bundle image restoration as an adaptive, context-aware decision process rather than a single fixed image-to-image model. GAME provides a single-entry point while dynamically routing each input to appropriate restoration experts according to the inferred imaging context. As shown in Supplementary Fig. 2, GAME follows a structure-to-detail workflow. First, it performs preliminary structural restoration to suppress honeycomb artifacts and recover reliable global morphology from the raw fiber-bundle image. Second, a VLM agent analyzes this preliminary restoration, infers the imaging context and formulates a reconstruction plan for expert selection. Third, the selected experts further refine local biological details, after which GAME evaluates the reliability of candidate outputs and updates its routing memory to improve subsequent decisions. This design enables GAME to achieve broad biomedical coverage while adapting to content- and wavelength-dependent degradation, without requiring prior knowledge of the sample type or imaging condition.

**Preliminary structural restoration.** Raw fiber-bundle images are often too degraded for reliable visual classification. Severe honeycomb artifacts, inter-core crosstalk and wavelength-dependent blur can obscure biological morphology, causing the VLM agent to misinterpret the imaging context or route the image to an unsuitable expert model. Therefore, before agent-based planning, GAME applies a unified Structural Restoration Model (SRM) to generate a preliminary restored image $I_{\text{pre}}$ from the raw image $I_{\text{in}}$. SRM is not intended to generate the final high-fidelity reconstruction. Instead, it provides a conservative structural estimate that suppresses dominant honeycomb artifacts and preserves reliable global morphology for subsequent VLM reasoning and expert refinement.

Directly learning a mapping from raw fiber-bundle images to clean references is challenging due to the substantial distribution gap between the two domains. We address this by formulating SRM as a decoupled residual diffusion model adapted from our recently proposed network [49], operating within a fixed noise-carrying domain (Supplementary Fig. 16). In SRM, controlled Gaussian noise is used as a fixed carrier to reduce the distribution gap between degraded and clean images, while the model learns to predict the deterministic degradation residual within this noise-carrying domain. This design constrains artifact removal to the degradation residual in a shared noise-conditioned space, thereby reducing the risk of hallucinating fine biological structures. For each training pair $(I_{\text{in}}, I_0)$, where $I_0$ is the clean reference or ground-truth image, we define the degradation residual as $I_{\text{res}} = I_{\text{in}} - I_0$. This residual $I_{\text{res}}$ captures the pixel-wise discrepancy introduced by fiber-bundle transmission. During training, the clean reference is first lifted to a fixed noise-carrying domain $(I_0 \rightarrow I_0 + \lambda_{\text{g}}\epsilon)$, and the degraded residual $I_{\text{res}}$ is then progressively injected:

$$I_t = I_0 + \lambda_{\text{g}}\epsilon + \bar{\alpha}_t I_{\text{res}}, \quad \epsilon \sim N(\mathbf{0}, \boldsymbol{I}) \tag{1}$$

where $\lambda_{\text{g}}$ denotes the fixed Gaussian noise level, $t$ denotes the timestep, and $\bar{a}_t = \prod_{i=1}^{t} a_i$ controls the residual injection schedule with $\bar{a}_0 = 0$, and $\bar{a}_T = 1$. At the final timestep, $I_T = I_0 + \lambda_{\text{g}}\epsilon + I_{\text{res}}$, which corresponds to a noise-carrying version of the raw fiber-bundle image. Here, $T$ is the number of sampling steps, and we set $T = 1000$ and $\lambda_{\mathrm{g}} = 2$ in our experiments. During reverse sampling, SRM starts from $I_T$ and removes the degraded residual while remaining in the same fixed noise-carrying domain. At each timestep, a time-conditioned residual predictor estimates:

$$\hat{I}_{\mathrm{res}}^{t} = R_{\theta}\left(I_t, t, I_{\mathrm{in}}\right) \tag{2}$$

and the diffusion state is updated as:

$$I_{t\text{-}1} = I_t - a_t \hat{I}_{\mathrm{res}}^{t}, \quad t = T, \ldots, 1 \tag{3}$$

where $R_{\theta}$ is a time-conditioned residual predictor implemented with a standard U-Net (Supplementary Fig. 16). The current diffusion state $I_t$ and the raw input image $I_{\mathrm{in}}$ are used as image conditions, while the timestep $t$ is encoded to guide step-wise residual removal. In our implementation, the U-Net uses a base channel number of 64 and channel multipliers of (1,2,4,8). SRM was trained with residual supervision:

$$\mathrm{L}_{\mathrm{SRM}} = \mathrm{E}_{I_{\mathrm{in}}, I_0, t, \epsilon}\left[\left\| I_{\mathrm{res}} - R_{\theta}\left(I_t, t, I_{\mathrm{in}}\right)\right\|_2^2\right]. \tag{4}$$

By predicting $I_{\mathrm{res}}$ instead of directly generating $I_0$, SRM constrains the preliminary restoration focused on artifact suppression and global structural recovery. During inference, SRM progressively subtracts the predicted residual in the fixed noise-carrying space, followed by removal of the known fixed Gaussian noise component $\lambda_{\mathrm{g}}\epsilon$ to obtain $I_{\mathrm{pre}}$. This preliminary result serves as a conservative structural prior for downstream VLM agent planning and expert refinement.

**VLM-agent-guided reconstruction planning and expert routing.** The VLM agent is the reasoning core of GAME. It interprets the preliminary restoration $I_{\mathrm{pre}}$, infers the biological context, and generates a restoration plan that determines which expert model or models should be active for detail refinement. This module consists of three components: visual perception, confidence-gated expert routing and user-controllable planning.

***(i) Visual perception***. Given $I_{\mathrm{pre}}$, the VLM agent uses the Qwen3.6-Flash API as the visual reasoning backbone to infer the biological context of the image. It returns a structured output consisting of a brief morphological description and confidence score $p_c$ over four predefined categories $C_{\mathrm{p}} = \{\text{cell, mouse, human, unknown}\}$ (Supplementary Fig. 2). Here, $p_c$ denotes the confidence score assigned to category $c \in C_{\mathrm{p}}$. These categories were chosen to cover the major biomedical scenarios in our dataset while retaining an *unknown* class for ambiguous or out-of-distribution inputs.

***(ii) Confidence-gated expert routing***. After obtaining $c$ and the corresponding confidence score $p_c$, the agent employs a confidence-gated routing policy to generate an expert restoration plan. The plan routes confidently recognized inputs to a single domain-specific expert, while activating multiple complementary experts for ambiguous inputs to improve robustness. Specifically, GAME maintains an expert pool $\varepsilon = \{E_{\text{cell}}, E_{\text{mouse}}, E_{\text{human}}, E_{\text{unknown}}\}$, where $E_{\text{cell}}, E_{\text{mouse}}$, and $E_{\text{human}}$ are domain-specific experts trained on corresponding biomedical data, and $E_{\text{unknown}}$ is a generalist fallback for out-of-distribution inputs (the concrete instantiation of these experts is described in the next section). Each perceptual category $c \in C_{\text{p}}$ is paired with its namesake expert $E_{\text{c}}$ (i.e., cell $\rightarrow E_{\text{cell}}$, mouse $\rightarrow E_{\text{mouse}}$, human $\rightarrow E_{\text{human}}$, unknown $\rightarrow E_{\text{unknown}}$); thus, selecting categories is equivalent to selecting experts. The agent identifies the most confident category $c^* = \arg\max_{c \in C_{\text{p}}} p_c$ and compares its confidence score against a threshold $\theta_{\text{conf}} = 0.75$. If $p_{c^*} \geq \theta_{\text{conf}}$, the input image is regarded as confidently recognized and agent activates only the matched expert:

$$C_{\text{exp}} = \{E_{c^*}\}, \quad \text{if } p_{c^*} \geq \theta_{\text{conf}}, \tag{5}$$

where $C_{\text{exp}}$ denotes the activated expert set. Conversely, if $p_{c^*} < \theta_{\text{conf}}$, the input image is regarded as category ambiguous or potentially out-of-domain, and the agent activates the two experts with the highest-confidence categories:

$$C_{\text{exp}} = \{E_c \mid c \in \text{Top2}_{c \in C_p}(p_c)\}, \quad \text{if } p_{\text{c}^*} < \theta_{\text{conf}}, \tag{6}$$

where $\text{Top2}_{c \in C_p}(p_c)$ returns the two categories with the highest confidence scores. The obtained $C_{\text{exp}}$ serves as the restoration plan to route appropriate experts for subsequent restoration.

***(iii) Optional user-controllable planning configuration***. The above pipeline operates fully automatically by default (Supplementary Fig. 17a). However, GAME also accepts optional user inputs to override or assist expert routing when prior knowledge is available. Users may provide a free-text hint (e.g., anatomical region or imaging condition or clinical context) to refine VLM inference (Supplementary Fig. 17b), or directly specify the desired expert to bypass perception entirely (Supplementary Fig. 17c). This flexibility accommodates scenarios where domain expertise or metadata can further reduce routing uncertainty, but is not required for standard operation.

**Expert restoration models for detail refinement.** Following the restoration plan generated by the VLM agent, GAME executes each activated expert $E_{\text{k}} \in C_{\text{exp}}$ as an Expert Restoration Model (ERM) for domain-adapted detail refinement. This section describes the concrete instantiation of the routing-level experts introduced above. For the *k*-th ERM, it takes the raw fiber-bundle image $I_{\text{in}}$ together with the structural prior $I_{\text{pre}}$ produced by SRM, and generates one expert-specific restoration candidate $I^{k}_{\text{refine}}$. When multiple experts are activated, GAME obtains multiple candidate restorations at this stage. These candidates are subsequently evaluated by the reliability reflection module to determine the final restored image.

The expert pool $\varepsilon = \{E_{\text{cell}}, E_{\text{mouse}}, E_{\text{human}}, E_{\text{unknown}}\}$ is instantiated as four ERMs. $E_{\text{cell}}$, $E_{\text{mouse}}$ and $E_{\text{human}}$ are domain-specific ERMs trained for cellular, mouse-tissue and human-tissue images, respectively, and $E_{\text{unknown}}$ is a generalist ERM for uncertain or out-of-distribution inputs. Additional ERMs can be incorporated by extending the expert pool $\varepsilon$ and updating the category-to-expert mapping, without modifying the overall reconstruction pipeline (Supplementary Fig. 3).

Recovering fine biological details from degraded fiber-bundle images requires both accurate artifact removal and preservation of global morphology. ERM addresses this requirement through two key designs: a hybrid residual-noise diffusion process that balances structural fidelity with generative flexibility, and an SRM-guided feature-fusion layer that anchors detail refinement to the preliminary structural restoration (Supplementary Fig. 18).

***(i) Hybrid residual-noise diffusion.*** Following Resfusion [50], we formulate a mixed residual-noise forward process for ERM, in which the clean image is progressively mixed with a residual term and Gaussian noise:

$$I_t = \sqrt{\bar{\mu}_t} I_0 + \left(1 - \sqrt{\bar{\mu}_t}\right) I_{\text{res}} + \sqrt{1 - \bar{\mu}_t}\,\epsilon, \quad \epsilon \sim \mathrm{N}(\mathbf{0}, \boldsymbol{I}), \tag{7}$$

where $\bar{\mu}_t = \prod_{i=1}^{t} \mu_i$ is the cumulative diffusion schedule, $t \in [0, T']$ denotes timestep. We use a truncated sampling steps with $T' = 5$ following Resfusion [50]. $\epsilon$ denotes the Gaussian noise, and $I_{\text{res}}$ denotes the residual between $I_{\text{in}}$ and $I_0$. As $t$ increases, $\bar{\mu}_t$ decreases from $\bar{\mu}_0 = 1$, progressively increasing the contributions of the residual and noise terms. In this formulation, the residual term acts as a deterministic correction signal that preserves structural fidelity and guides fiber-bundle degradation removal, whereas the noise term introduces stochastic generative flexibility for recovering fine structures. By reparameterizing the forward process in Eq. (7), the ERM is trained to predict a hybrid residual-noise target:

$$\eta_t = \epsilon + \frac{\left(1 - \sqrt{\mu_t}\right)\sqrt{1 - \bar{\mu}_t}}{1 - \mu_t} I_{\text{res}} \tag{8}$$

***(ii) SRM-guided feature fusion and training objective.*** The generative refinement described above may still introduce details that are inconsistent with the underlying sample morphology. We therefore use the SRM branch, which conservatively preserves low-frequency morphology, as a structural prior to constrain ERM refinement. Because the SRM and ERMs share the same time-conditioned U-Net backbone, we fuse their intermediate features at the end of the middle block, where both networks encode structural information at a comparable abstraction level. Specifically, let $F_{\mathrm{SRM}}$ and $F_{\mathrm{ERM}}^{k}$ denote the intermediate feature extracted from last layer of the middle block of SRM and the *k*-th ERM, respectively. A lightweight projection layer aligns the SRM feature to the ERM feature space before fusion:

$$F_{\mathrm{fused}}^{k} = F_{\mathrm{ERM}}^{k} + \gamma_k \, \mathrm{f}_{\phi_k}\left(F_{\mathrm{SRM}}\right) \tag{9}$$

where $\mathrm{f}_{\phi_k}(\cdot)$ is a learnable projection layer for feature-space alignment, and $\gamma_k \in [0,1]$ is a learnable gating weight that controls the contribution of SRM feature. The fused feature is then processed by subsequent ERM layers. This design allows the ERM to restore fine details while remaining anchored to the globally reliable structure estimated by SRM. This fusion module introduces only a small number of additional parameters (~0.1M), resulting in approximately 36.5M parameters for each ERM.

The *k*-th ERM is trained to predict the hybrid residual-noise target in Eq. (8) by minimizing:

$$\mathrm{L}_{\mathrm{ERM}}^{k} = \mathrm{E}\left[\left\|\eta_t - \eta_{\theta_k}\left(I_t, t, I_{\mathrm{in}}; F_{\mathrm{SRM}}\right)\right\|_2^2\right] \tag{10}$$

where $\eta_{\theta_k}$ denotes the residual-noise prediction network of the *k*-th ERM.

***(iii) ERM inference and candidate generation.*** During inference, ERM follows the Resfusion reverse process [50] to generate an expert-refined candidate. Unlike standard diffusion models that start from pure Gaussian noise, the reverse chain is initialized from a noisy degraded input. This initialization allows ERM to refine the degraded structure rather than reconstructing the image from scratch. Starting from the truncated step $t = T'$, the *k*-th ERM iteratively predicts the residual-noise term and updates the current estimate as:

$$I_{t\text{-}1} = \frac{1}{\sqrt{\mu_t}}\left(I_t - \frac{\beta_t}{\sqrt{1-\bar{\mu}_t}}\eta_{\theta_k}\left(I_t, t, I_{\mathrm{in}}; F_{\mathrm{SRM}}\right)\right) + \sqrt{\tilde{\beta}_t}\,\varepsilon_t, \quad \varepsilon_t \sim \mathrm{N}(\mathbf{0}, \boldsymbol{I}). \tag{11}$$

Here, $\beta_t = 1 - \mu_t$, $\tilde{\beta}_t = \frac{1-\bar{\mu}_{t-1}}{1-\bar{\mu}_t}\beta_t$ is the reverse variance. $\varepsilon_t$ denotes the step-wise Gaussian noise used in reverse sampling. Through this iterative update, ERM subtracts the predicted residual-noise term at each step, jointly reducing the weighted residual and noise components to suppress fiber-bundle degradations and recover fine biological structures. After the last reverse step, the output of the *k*-th ERM is taken as an expert-refined candidate: $I_{\mathrm{refine}}^{k} = I_0^{k}$. This candidate is passed to the subsequent reliability reflection stage for candidate verification and final output selection.

**Agent-guided reliability reflection and routing memory update**

***(i) Reliability reflection.*** After expert restoration, GAME performs reliability reflection to select the final output from the candidate set $I_{\mathrm{refine}}^{k}$, where $k \in C_{\exp}$. Although ERMs are designed to enhance domain-specific fine details, an unsuitable expert may introduce details that are inconsistent with the global morphology. Therefore, this step uses the SRM preliminary restoration $I_{\text{pre}}$ as a conservative structural anchor to reflect on whether each expert-refined candidate remains structurally reliable. An expert-refined candidate is considered more reliable if its low-frequency structure remains consistent with $I_{\text{pre}}$. To evaluate this structural consistency, we apply a Gaussian blur operator $G(\cdot)$ to both $I_{\text{pre}}$ and each expert-refined candidate $I_{\text{refine}}^{k}$. This operation suppresses high-frequency detail differences and retains low-frequency information, allowing the agent to assess whether the global structure of $I_{\text{refine}}^{k}$ has deviated from the $I_{\text{pre}}$. The reliability score $S_{\text{R}}^{k}$ is defined as:

$$S_{\text{R}}^{k} = \text{SSIM}\left(G\left(I_{\text{pre}}\right), G\left(I_{\text{refine}}^{k}\right)\right) \tag{12}$$

where SSIM denotes the structural similarity index measure. A higher $S_{\text{R}}^{k}$ indicates a stronger structural agreement with the SRM output. GAME then selects the candidate with the highest structural reliability score as the final reconstruction output:

$$k^{*} = \arg\max_{k \in C_{\exp}} S_{\text{R}}^{k} \tag{13}$$

$$\hat{I} = I_{\text{refine}}^{k^{*}} \tag{14}$$

where $\hat{I}$ is the final output. In this way, reliability reflection serves as a candidate-selection mechanism grounded in structural consistency, allowing GAME to accept expert-generated fine details only when they remain consistent with the reliable global morphology preserved by SRM.

***(ii) Rule-informed self-evolution.*** VLM agent enables flexible expert routing by reasoning over the SRM-restored image $I_{\text{pre}}$ rather than the raw fiber-bundle measurement. However, visually ambiguous cases can still lead to incorrect expert selection, since different biomedical domains share similar low-frequency morphology and the wavelength-dependent degradation may alter apparent texture and contrast. Such routing errors are difficult to eliminate using a fixed prompt or static category descriptions alone. To improve routing robustness, GAME incorporates a rule-informed self-evolution mechanism that converts verified routing failures into explicit textual decision rules for future expert selection (Supplementary Fig. 2).

We define a routing failure as a case in which the top-1 confident expert identified by the VLM agent is inconsistent with the ground-truth expert label. To support rule-informed self-evolution, GAME maintains a dynamic routing memory bank that stores failure-derived textual decision rules. During training, if a routing failure is identified, GAME records the failure case and updates this dynamic routing memory bank. Specifically, for each verified failure case, GAME records the SRM-restored image $I_{\text{pre}}$, category confidence score $p_c$, the incorrectly selected expert, and the corrected expert label. These records are organized into a failure-analysis prompt and subsequently fed back to the VLM agent, prompting the VLM agent to diagnose why the original expert selection failed, identify the discriminative patterns that distinguish the incorrect and corrected label, and summarize these cues into compact textual routing rules. The summarized rules are then appended to our dynamic routing memory bank as reusable routing experience.

During subsequent inference, GAME retrieves the latest ten rules from the memory bank and inserts them into the VLM prompt additional reasoning context. In this way, the VLM agent performs routing not only based on the current SRM-restored image $I_{\text{pre}}$, but also with reference to recently accumulated failure-derived experience. This allows the agent to avoid previously observed routing mistakes when similar ambiguous patterns reappear. The process updates only the textual memory bank and does not modify the SRM or ERM parameters.

## Training details

**Data augmentation.** Raw degraded inputs and their paired ground-truth images were processed at the original resolution of $640 \times 512$ pixels. To improve robustness to spatial variations during fiber-bundle imaging, we applied a data augmentation pipeline during training. For each paired sample, affine augmentation was applied with a probability of 0.5, including random translation within $\pm 20\%$ along both spatial axes and random rotation within $[-45^{\circ}, 45^{\circ}]$. Random horizontal and vertical flips were also applied independently, each with a probability of 0.5. To ensure a fair comparison, all the models used the same data augmentation pipeline for training.

**Structural restoration model.** To achieve preliminary global structural restoration, the SRM was trained on the complete Fiber 2 dataset, using an 8:2 split of training and validation, corresponding to 59,561 paired training images and 14,891 validation images. The network was optimized using the Adam optimizer for 500,000 steps with an initial learning rate of $2\times10^{-5}$. An exponential moving average (EMA) with a decay factor of 0.995 was applied to the model weights. Training was distributed across six NVIDIA L40S GPUs (48 GB VRAM each) with a local batch size of one per GPU. We employed the same configuration to train dedicated SRMs for Fiber 1 and Fiber 3. Notably, the final outputs reported in this study were generated by the ERMs, with the SRM only providing structural prior features for the ERMs or preliminary reconstructions for VLM reasoning.

**Expert restoration models.** A series of ERMs were trained to perform domain-specific detail restoration. To construct the agent-guided framework, we trained four ERMs on the Fiber 2 dataset, comprising a Cell ERM, a Mouse ERM, a Human ERM, and a Generalist ERM. These expert networks correspond one-to-one with the candidate categories evaluated during the VLM routing phase, allowing the VLM's routing decisions to select designated experts. Specifically, the Cell, Mouse, and Human ERMs were trained to process inputs assigned to their respective biological domains, while the Generalist ERM was dedicated to handling the unknown category. For each expert model, the training and validation datasets were partitioned using an 8:2 split. The Cell ERM was trained on 14,720 paired images and validated on 3,680 pairs; the Mouse ERM used 5,488 training and 1,372 validation pairs; the Human ERM was trained with 2,578 training and 645 validation pairs; and the Generalist ERM was trained on the complete Fiber 2 dataset, which contains 59,561 training and 14,891 validation pairs.

Additional ERMs were trained to evaluate restoration performance under varying wavelengths and FOVs. For the wavelength-dependent evaluation (Fig. 2), three fiber-specific ERMs were trained on the Fiber 1, Fiber 2, and Fiber 3 datasets, respectively. Each dataset contained mouse images acquired at five wavelengths and was split into 3,600 training and 900 validation pairs. For FOV-specific evaluation, we used the Fiber 2 dataset containing six FOVs with magnification factors from 1.0 to 2.0 at 0.2 increments, resulting in 7,776 training and 1,944 validation pairs.

All ERMs were optimized using AdamW for 200 epochs, except for the Generalist ERM, which converged within 50 epochs, with an initial learning rate of $1\times10^{-4}$. Training was distributed across NVIDIA L40S GPUs (48 GB VRAM each) with a local batch size of 4 per GPU and a gradient accumulation of 4.

**Comparison models.** To ensure a fair comparison, U-Net [33] and UniFMIR [32] were trained using the complete datasets acquired from all three fiber bundles. This allowed the comparison models to be exposed to a data scale comparable to that used for GAME training. The same data preprocessing and augmentation strategies were applied to all comparison models. During evaluation, U-Net, UniFMIR, and GAME were tested on the same test images, and their reconstruction performance was compared using PSNR, SSIM, and LPIPS.

## Evaluation metrics

We assessed image restoration quality using the Peak Signal-to-Noise Ratio (PSNR), Structural Similarity Index Measure (SSIM), and Learned Perceptual Image Patch Similarity (LPIPS). To quantitatively validate the super-resolution performance and the elimination of honeycomb artifacts, we analyzed the Full Width at Half Maximum (FWHM) of the resolved microvasculature. Specifically, for each experimental condition in Fig. 3b,c, cross-sectional intensity profiles from 30 distinct blood vessels were extracted and fitted to 1D Gaussian distributions to analytically determine the FWHM. The measured FWHM was served as the feature size ($S_F$) to analyze the resolution limit of our restoration model.

## Animal ethics statement

All animal experiments were approved by the Materials Innovation Institute for Life Sciences and Energy (MILES), The University of Hong Kong-SIRI (Protocol No. 21-2510-2). Six-week-old BALB/c female mice were used for in vivo imaging. Mice were selected randomly from cages for experiments.

**In vivo AI-powered flexible fiber-bundle endoscopy**

For in vivo NIR-I and NIR-II fluorescence endoscopy, an FBE was fabricated using Fiber 2. A custom miniature lens with a 60° field of view and a focal length of ~0.5 mm was integrated at the distal tip of the fiber bundle using a 3D-printed threaded holder, which was further secured with UV-curable adhesive. An 808-nm laser was delivered to the target region through an auxiliary illumination fiber positioned adjacent to the endoscope for ICG excitation. The emitted fluorescence was collected by the miniature lens, transmitted through the fiber bundle, and imaged using a wide-field microscope equipped with a 10× objective and a 200-mm tube lens. Fluorescence signals were recorded using an InGaAs camera. Mice were anesthetized using a rodent anesthesia machine with 2 L/min $O_2$ gas mixed with 3% isoflurane during ICG administration and imaging.

For NIR-II fluorescence imaging of the intestine and liver as shown in Figure 4c and 4d, mice received a retro-orbital intravenous injection of ICG solution at 2 mg/mL, with an injection volume of 150 µL per mouse. The intestine and liver were imaged after abdominal exposure at 2 h and 3-4 h after injection, respectively. Fluorescence emission was collected through a 1100-nm long-pass filter for intestine and a 1000-nm long-pass filter for liver imaging, using our fiber-bundle endoscope. The exposure time was 50 ms and 25 ms for intestine and liver imaging, respectively. For each organ, raw fiber-bundle endoscopic images and corresponding wide-field reference images were acquired under similar imaging conditions. The raw endoscopic images were used as inputs for AI-based image restoration.

For ureter imaging shown in Figure 4e, we followed a procedure used for ICG-guided ureter localization in human surgery [40]. A sterilized capillary tube with an outer diameter of 0.52 mm was inserted into the mouse urethra. The other end of the capillary tube was connected to an insulin needle. An initial 100 µL injection of 2 mg/mL ICG solution was administered through the capillary tube. Ten minutes later, the remaining 200 µL of ICG solution was injected into the mouse through the tube. The mice were then dissected, and the ureters and kidneys were exposed for fluorescence imaging. ICG fluorescence emission was collected after passing through an 830-nm long-pass filter. The exposure time was 30 ms.

For the NIR-II fluorescence imaging of lymph nodes and lymphatic vessels shown in Figure 4f, 80 µL of ICG solution at a concentration of 2 mg/mL was administered subcutaneously near the base of the mouse tail as two 40-µL aliquots. Thirty minutes after injection, fluorescence imaging of the mouse lymph nodes and lymphatic vessels was performed. The emitted fluorescence signal was collected after passing through a 1000-nm long-pass filter, with an exposure time of 50 ms.

## Acknowledgments

This study was supported by the National Natural Science Foundation of China (Project No. T2522030, 62306253), the Early Career Scheme (RGC No. 27204623, 27207025) from the Research Grants Council of Hong Kong SAR, the NSFC/RGC Collaborative Research Scheme (CRS_HKU703/25), start-up funding from Materials Innovation Institute for Life Sciences and Energy (MILES), HKU-SIRI in Shenzhen, and the Guangdong Natural Science Fund-General Programme (Project No. 2024A1515010233).

## Author contributions

F.W., H.D., and L.Q. conceived the project and designed the experiments. F.W. designed the fiber-bundle endoscope. L.Q. and Y.S. designed the GAME framework. Y.S. implemented the GAME framework and performed the core experiments. Y.L. was responsible for constructing the optical system, collecting data, and performing the initial AI network verification. S.X. and Y.L. conducted the primary animal experiments. W.J.L., Y.C., D.X., Z.W., I.Y.W., and S.Y.L. contributed to data organization and assisted with the biomedical experiments. Y.S., F.W. and L.Q. wrote the manuscript. L.Q. and F.W. provided overall project guidance and supervision. All authors contributed to the general discussion and revision of the manuscript.

**Competing interests**

Authors declare no competing interests.

**Materials & Correspondence**

Correspondence and requests for materials should be addressed to F.W. (feifwang@hku.hk).

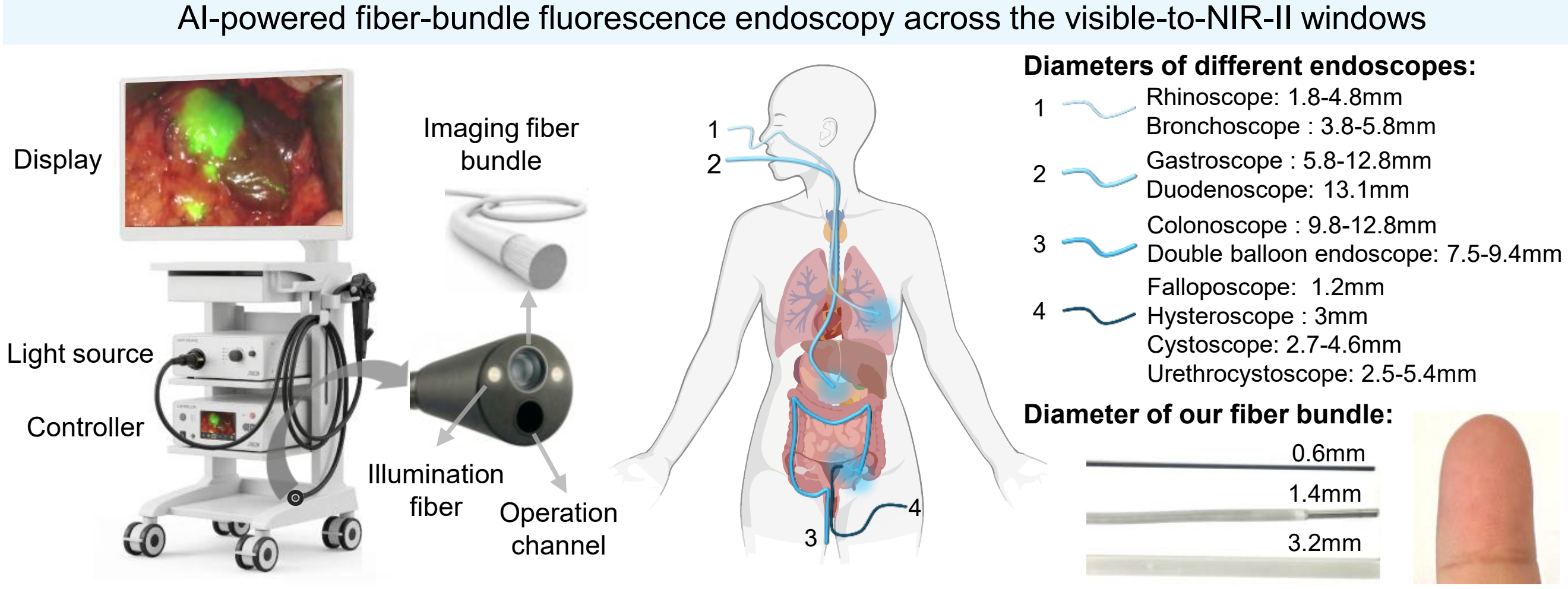
AI-powered fiber-bundle fluorescence endoscopy across the visible-to-NIR-II windows
Display
Light source
Controller
Imaging fiber bundle
Illumination fiber
Operation channel
Diameters of different endoscopes:
Rhinoscope: 1.8-4.8mm
Bronchoscope : 3.8-5.8mm
Gastroscope : 5.8-12.8mm
Duodenoscope: 13.1mm
Colonoscope : 9.8-12.8mm
Double balloon endoscope: 7.5-9.4mm
Falloposcope: 1.2mm
Hysteroscope : 3mm
Cystoscope: 2.7-4.6mm
Urethrocystoscope: 2.5-5.4mm
Diameter of our fiber bundle:
0.6mm
1.4mm
3.2mm

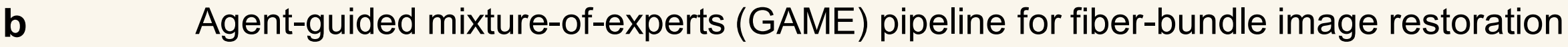
b
Agent-guided mixture-of-experts (GAME) pipeline for fiber-bundle image restoration

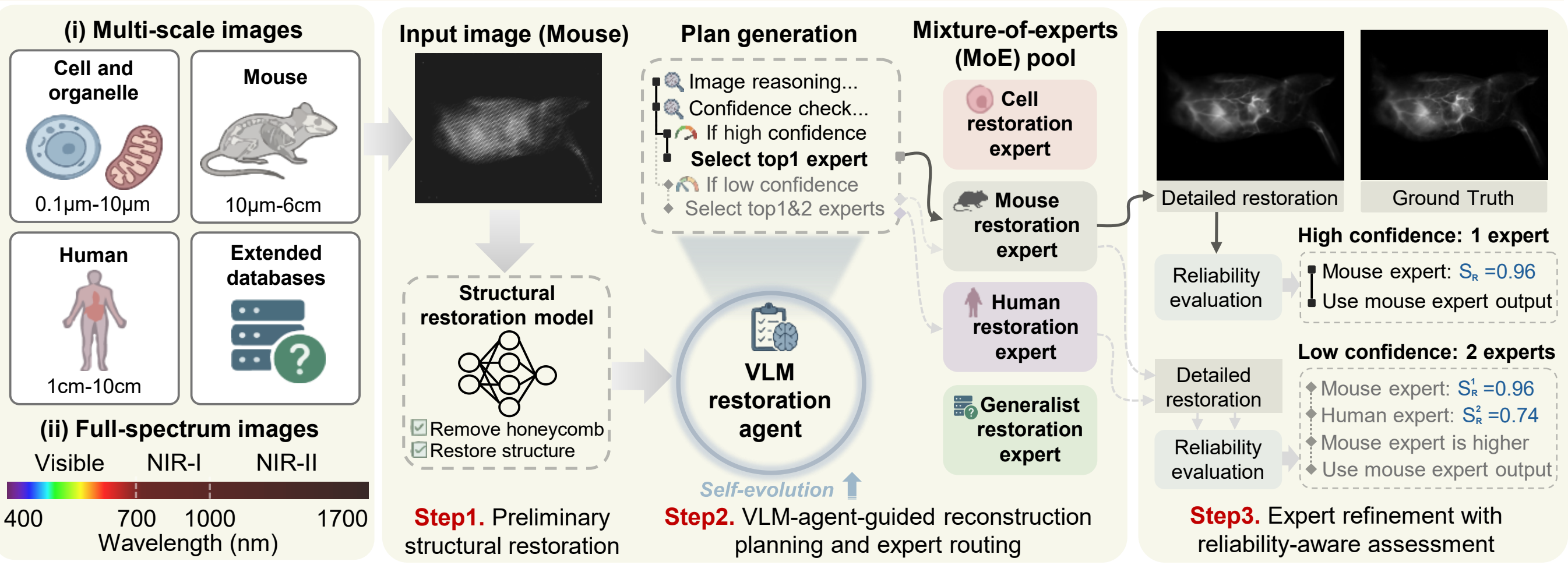
(i) Multi-scale images
Cell and organelle
0.1μm-10μm
Mouse
10μm-6cm
Human
1cm-10cm
Extended databases
(ii) Full-spectrum images
Visible NIR-I NIR-II
400 700 1000 1700
Wavelength (nm)
Input image (Mouse)
Structural restoration model
Remove honeycomb
Restore structure
Plan generation
Image reasoning...
Confidence check...
If high confidence
Select top1 expert
If low confidence
Select top1&2 experts
VLM restoration agent
Self-evolution
Mixture-of-experts (MoE) pool
Cell restoration expert
Mouse restoration expert
Human restoration expert
Generalist restoration expert
Detailed restoration
Ground Truth
High confidence: 1 expert
Reliability evaluation
Mouse expert: $S_R$ =0.96
Use mouse expert output
Low confidence: 2 experts
Detailed restoration
Reliability evaluation
Mouse expert: $S_R^1$ =0.96
Human expert: $S_R^2$ =0.74
Mouse expert is higher
Use mouse expert output
Step1. Preliminary structural restoration
Step2. VLM-agent-guided restoration planning and expert routing
Step3. Expert refinement with reliability-aware assessment

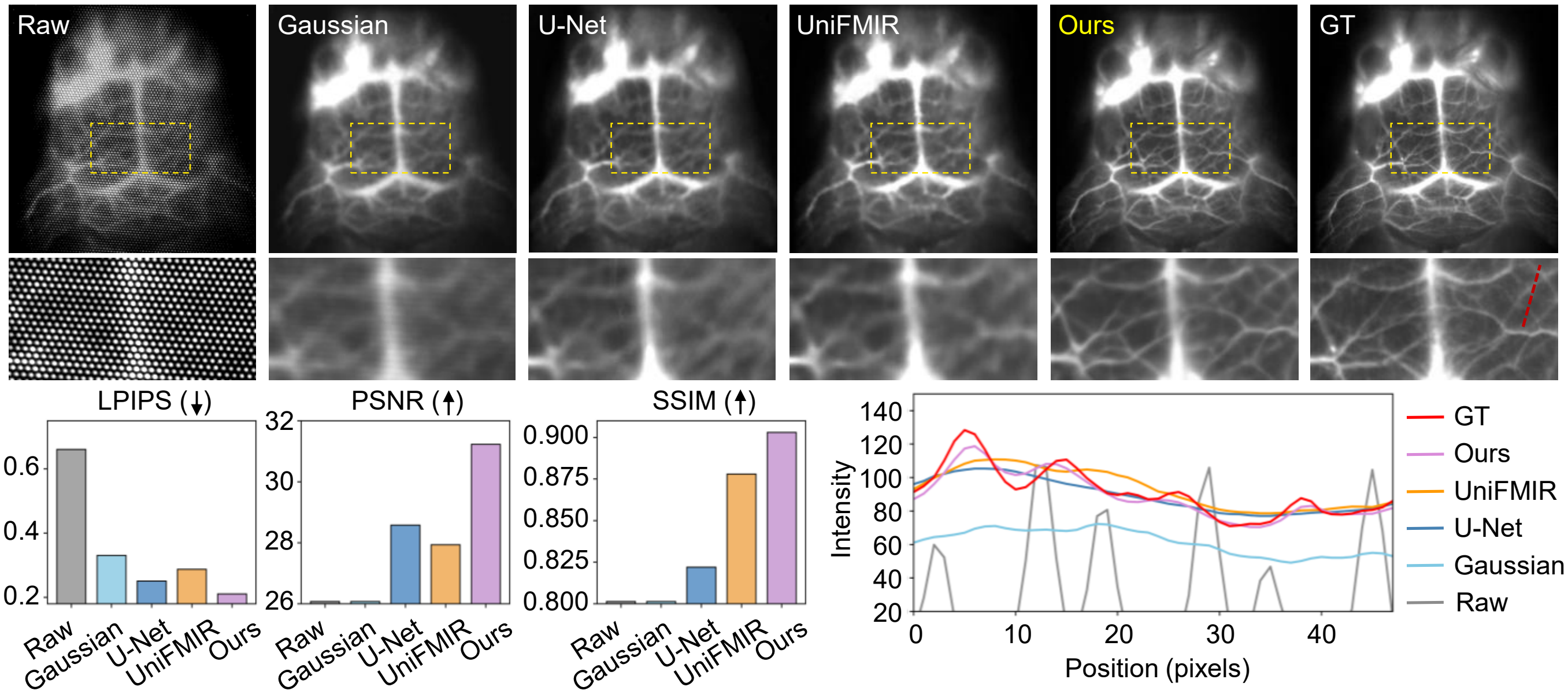
c
Image reconstruction results
Raw
Gaussian
U-Net
UniFMIR
Ours
GT
LPIPS (↓)
PSNR (↑)
SSIM (↑)
Raw
Gaussian
U-Net
UniFMIR
Ours
Intensity
Position (pixels)
GT
Ours
UniFMIR
U-Net
Gaussian
Raw

**Figure 1 | Flexible fiber-bundle endoscope across the visible-to-NIR-II window powered by an Agent-Guided Mixture-of-Experts (GAME) network.** (**a**) Schematic of the wide-spectrum (visible-to-NIR-II) fluorescence endoscopic platform, shown as an example, integrating an imaging fiber bundle, an illumination fiber, and an operation channel. The diameter comparison highlights the ultrathin profile of our fiber bundles (0.6, 1.4, and 3.2 mm) relative to standard clinical endoscopes, demonstrating their accessibility for access through confined human orifices. (**b**) The GAME computational framework. Left: The experimentally collected dataset encompasses 96,050 paired images of cells, mouse tissues, and human tissues, with feature sizes ranging from 0.1 μm to 10 cm and imaging wavelengths covering 485-1550 nm. Right: The three-step GAME image restoration workflow: Step 1, preliminary structural restoration to remove honeycomb artifacts; Step 2, VLM-based decision-making, which evaluates image type and confidence values to dynamically invoke either the highest-confidence restoration expert for high-confidence cases or the top two ranked restoration experts for low-confidence cases; Step 3, expert-specific detail restoration followed by reliability evaluation to generate the final output. For high-confidence cases, only one reconstruction and its corresponding reliability value are output, whereas for low-confidence cases, the reconstruction with the highest reliability value is selected as the final output. (**c**) Representative reconstructions of an in vivo image of the mouse head vasculature. Comparison with the ground truth (GT) demonstrates the superior structural fidelity and artifact suppression achieved by the GAME pipeline compared with a conventional Gaussian filter and other deep learning architectures, including U-Net and UniFMIR. The LPIPS, PSNR, and SSIM values were used to quantitatively compare the input and reconstructed images relative to the GT data. Intensity profiles measured along the red dotted line in each image indicate that our reconstruction matches the GT with high fidelity.

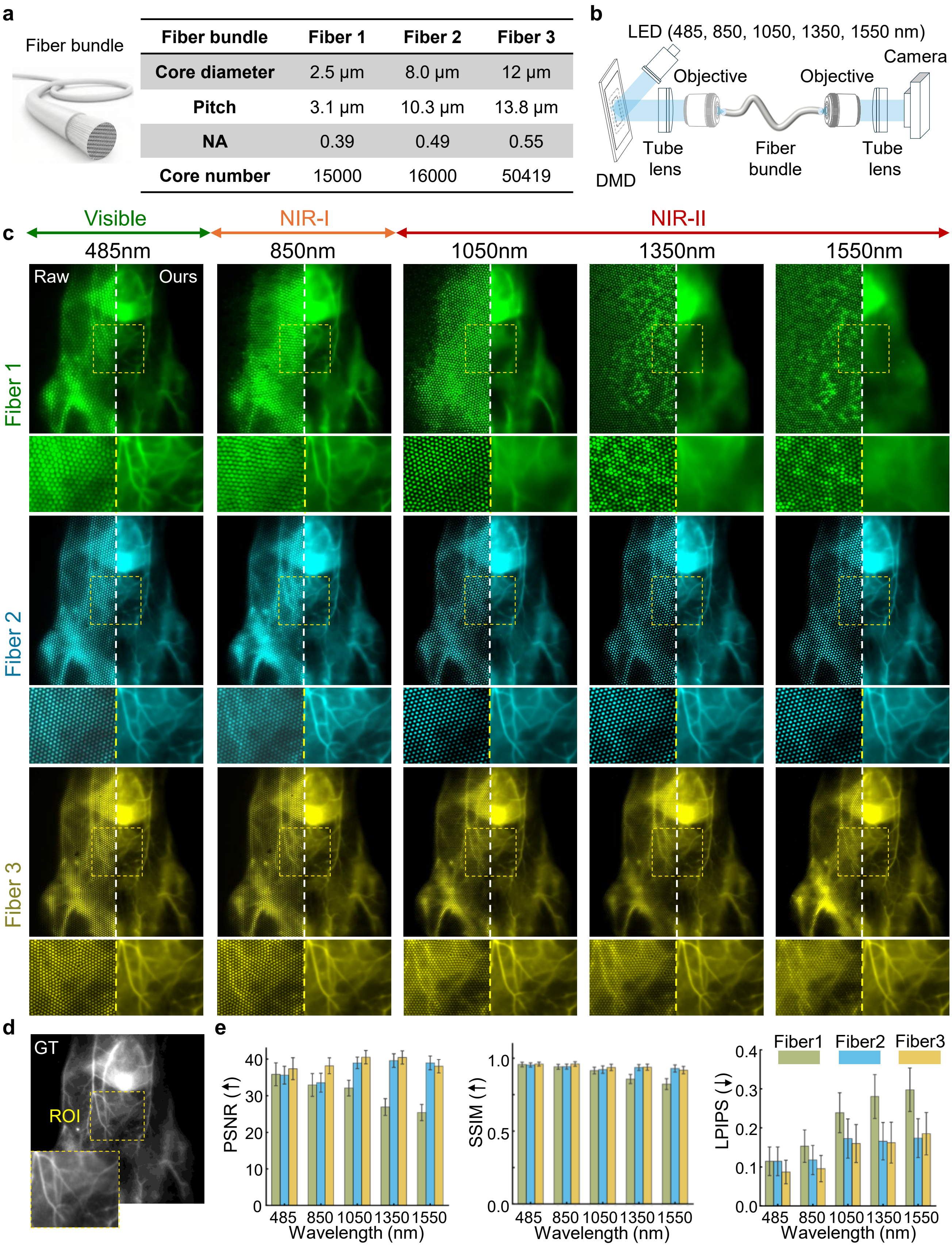

| Fiber bundle | Fiber 1 | Fiber 2 | Fiber 3 |
|---|---|---|---|
| Core diameter | 2.5 μm | 8.0 μm | 12 μm |
| Pitch | 3.1 μm | 10.3 μm | 13.8 μm |
| NA | 0.39 | 0.49 | 0.55 |
| Core number | 15000 | 16000 | 50419 |

**Figure 2 | Mitigation of inter-core crosstalk and GAME-based image reconstruction across the visible to NIR-II windows.** (**a**) Three fiber bundles (Fiber 1, Fiber 2, and Fiber 3), with different core diameters, core-to-core pitches, NAs, and total numbers of fiber cores, were used for imaging. (**b**) Schematic of the DMD-based imaging system for paired data acquisition. Ground-truth images, including images of cells, mouse tissues, and human tissues, were input into the DMD, which reflected LED illumination at 485, 850, 1050, 1350, or 1550 nm to mimic fluorescence imaging from the visible to the NIR-II window. The generated image was collected by a tube lens and then demagnified onto the distal end of the fiber bundle using an objective, while the transmitted image from the proximal end was imaged by a wide-field microscope equipped with an objective, a tube lens and a camera. (**c**) The influence of fiber-bundle configuration on image transmission and reconstruction by the GAME network across visible-to-NIR-II wavelengths. Each panel contrasts the raw honeycomb-degraded input on the left with the computationally restored output on the right. Dashed boxes indicate the magnified regions of interest (ROIs) displayed in the rows below. (**d**) The ground-truth image with the selected ROI corresponding to the images shown in **c**. (**e**) Quantitative evaluation of the restored images using PSNR, SSIM, and LPIPS metrics. Data are presented as the mean ± s.d. across the test dataset.

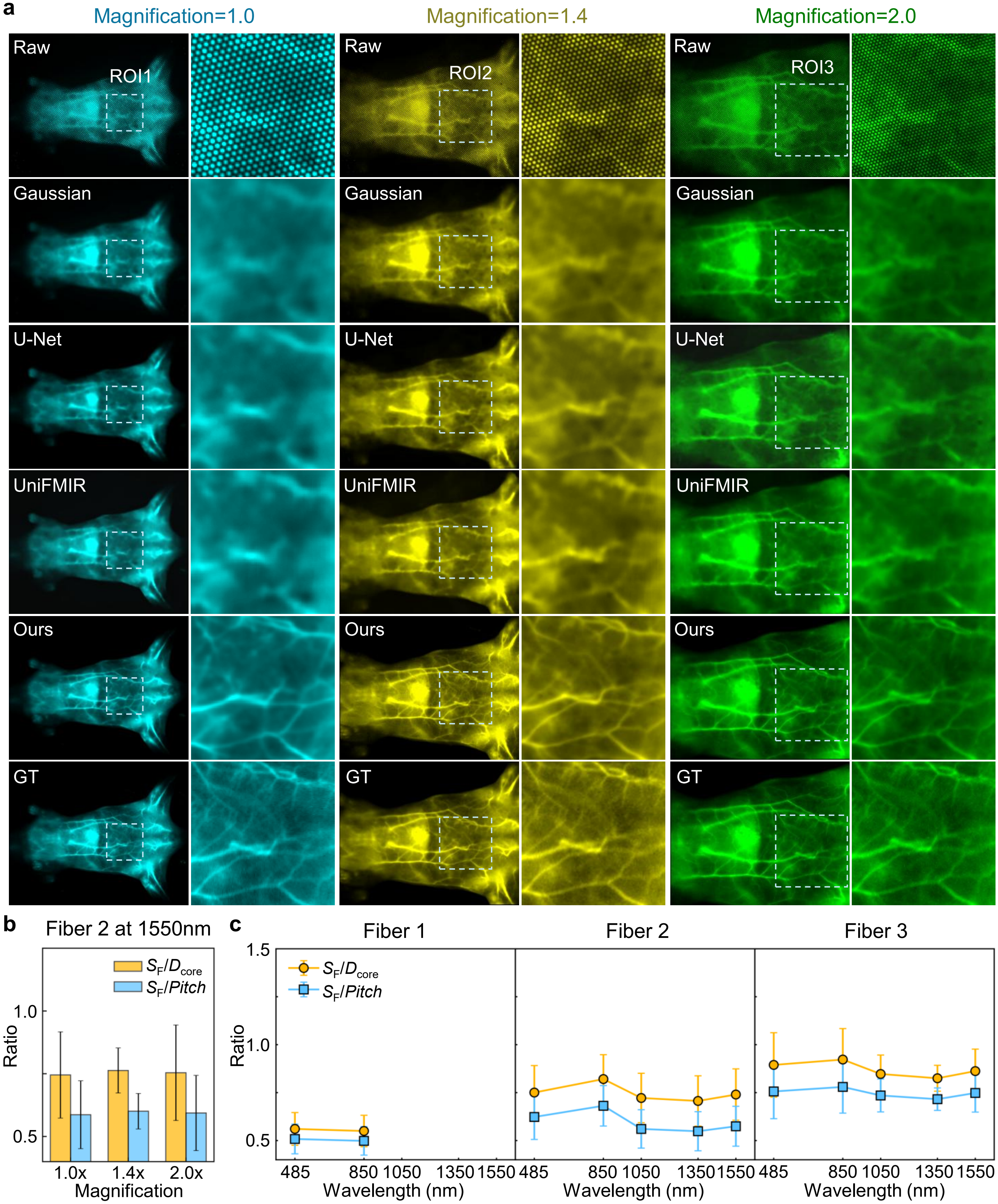
a
Magnification=1.0
Magnification=1.4
Magnification=2.0
Raw
ROI1
ROI2
ROI3
Gaussian
U-Net
UniFMIR
Ours
GT
b
Fiber 2 at 1550nm
$S_F/D_{core}$
$S_F/Pitch$
Ratio
1.0
0.5
1.0x
1.4x
2.0x
Magnification
c
Fiber 1
Fiber 2
Fiber 3
1.5
1.0
0.5
485
850
1050
1350
1550
Wavelength (nm)

**Figure 3 | Assessment of reconstruction resolution in fiber-bundle imaging.** (**a**) Representative reconstructions of mouse blood vessels acquired using Fiber 2 with Gaussian filtering, U-Net, UniFMIR [32], and our method at different magnifications. Images at different magnifications were generated by scaling the ground-truth image input to the DMD by magnification factors of 1.0×, 1.4×, and 2.0×. As the scaling factor increased, more fiber cores sampled the same vascular structure. Dashed white boxes indicate the magnified ROIs. A 1550-nm LED was used for illumination. (**b**) Quantitative comparison of $S_F/D_{core}$ and $S_F/Pitch$ for Fiber 2 at 1550 nm under different magnifications using GAME network, where $S_F$ denotes the FWHM of the smallest restorable blood vessels shown in **a**. (**c**) Quantitative analysis of $S_F/D_{core}$ and $S_F/Pitch$ for three fiber bundles across the visible to NIR-II windows. For **b** and **c**, data are presented as mean ± s.d. (n=30 independent blood vessels per condition).

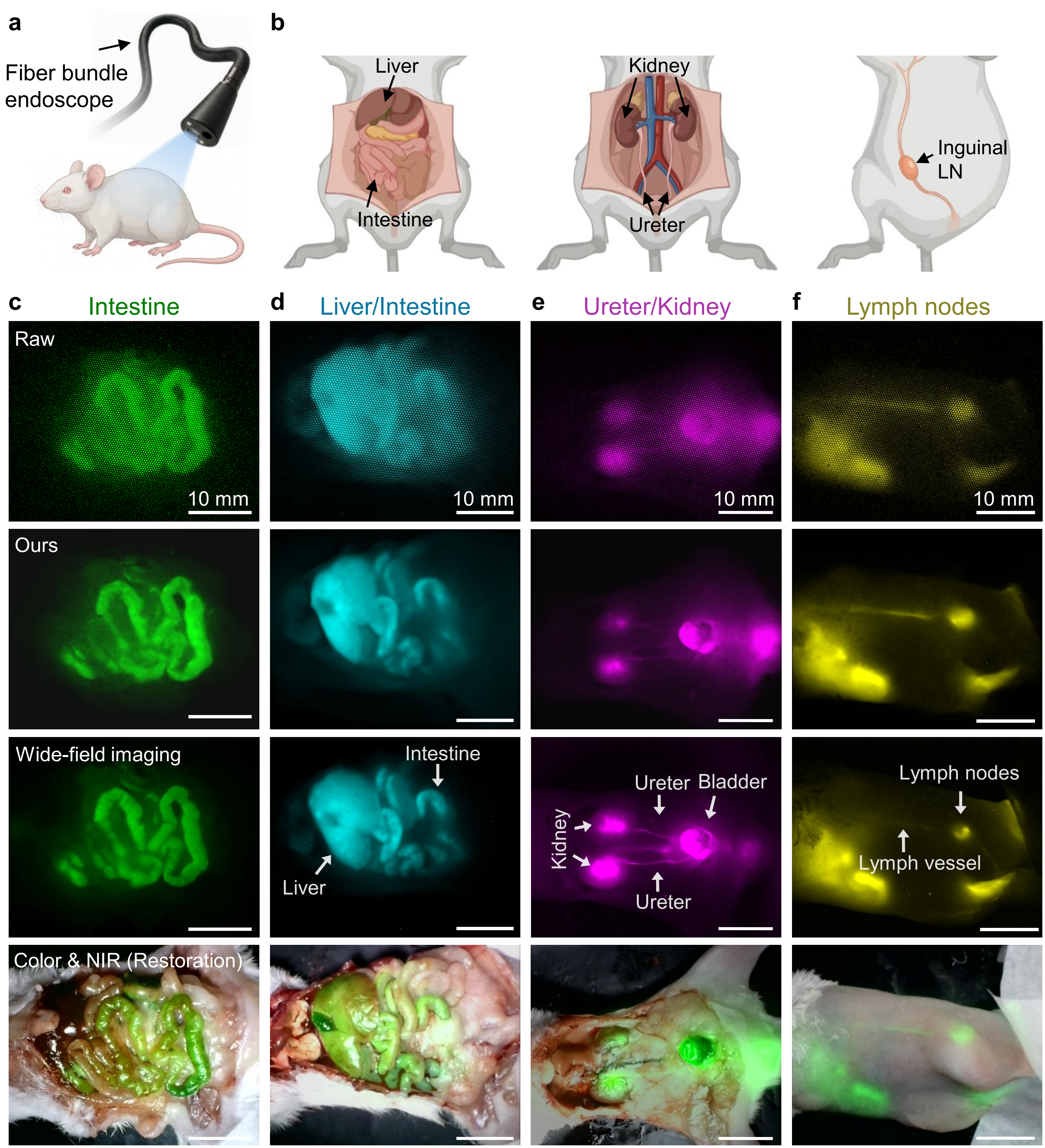

a
Fiber bundle
endoscope
b
Liver
Intestine
Kidney
Ureter
Inguinal
LN
c
Intestine
d
Liver/Intestine
e
Ureter/Kidney
f
Lymph nodes
Raw
10 mm
10 mm
10 mm
10 mm
Ours
Wide-field imaging
Intestine
Liver
Ureter
Bladder
Kidney
Ureter
Lymph nodes
Lymph vessel
Color & NIR (Restoration)

**Figure 4 | In vivo NIR-I and NIR-II AI-powered flexible fiber-bundle endoscopy.** (**a**) Schematic of in vivo NIR-I and NIR-II flexible fiber-bundle endoscopic imaging of a mouse. We installed a custom miniature lens with a 60° field of view at the distal tip of a fiber-bundle endoscope fabricated using Fiber 2. The proximal end of fiber bundle was coupled to a wide-field microscope equipped with a 10× objective and a tube lens for NIR-I and NIR-II fluorescence imaging. (**b**) Anatomical illustrations of three animal models for imaging of the liver, intestine, kidneys, ureters, lymph nodes, and lymphatic vessels. (**c**,**d**) Raw NIR-II endoscopic images, images reconstructed using GAME, wide-field images, and overlays of color and reconstructed images of the mouse intestine and liver. The intestine and liver were imaged after abdominal exposure at 2 h and 3-4 h post retro-orbital injection of ICG, respectively. Fluorescence was collected through a 1100-nm long-pass filter with a 50-ms exposure time in (**c**), and a 1000-nm long-pass filter with a 25-ms exposure time in (**d**). Raw endoscopic images and corresponding wide-field reference images were acquired under similar experimental conditions for further comparison. (**e**) The raw NIR-I endoscopic image, the reconstructed image, the wide-field image, and the overlay of color and reconstructed images of the mouse ureter. ICG solution was delivered through a sterilized capillary tube inserted into the mouse urethra. The ureters and kidneys were then exposed and imaged using an 830-nm long-pass filter with a 30-ms exposure time. (**f**) The raw NIR-II endoscopic image, the GAME-reconstructed image, the wide-field image, and the overlay of color and reconstructed images of lymph nodes and lymphatic vessels. ICG was injected subcutaneously near the base of the tail. Imaging was performed 30 min after injection using a 1000-nm long-pass filter with a 50-ms exposure time.

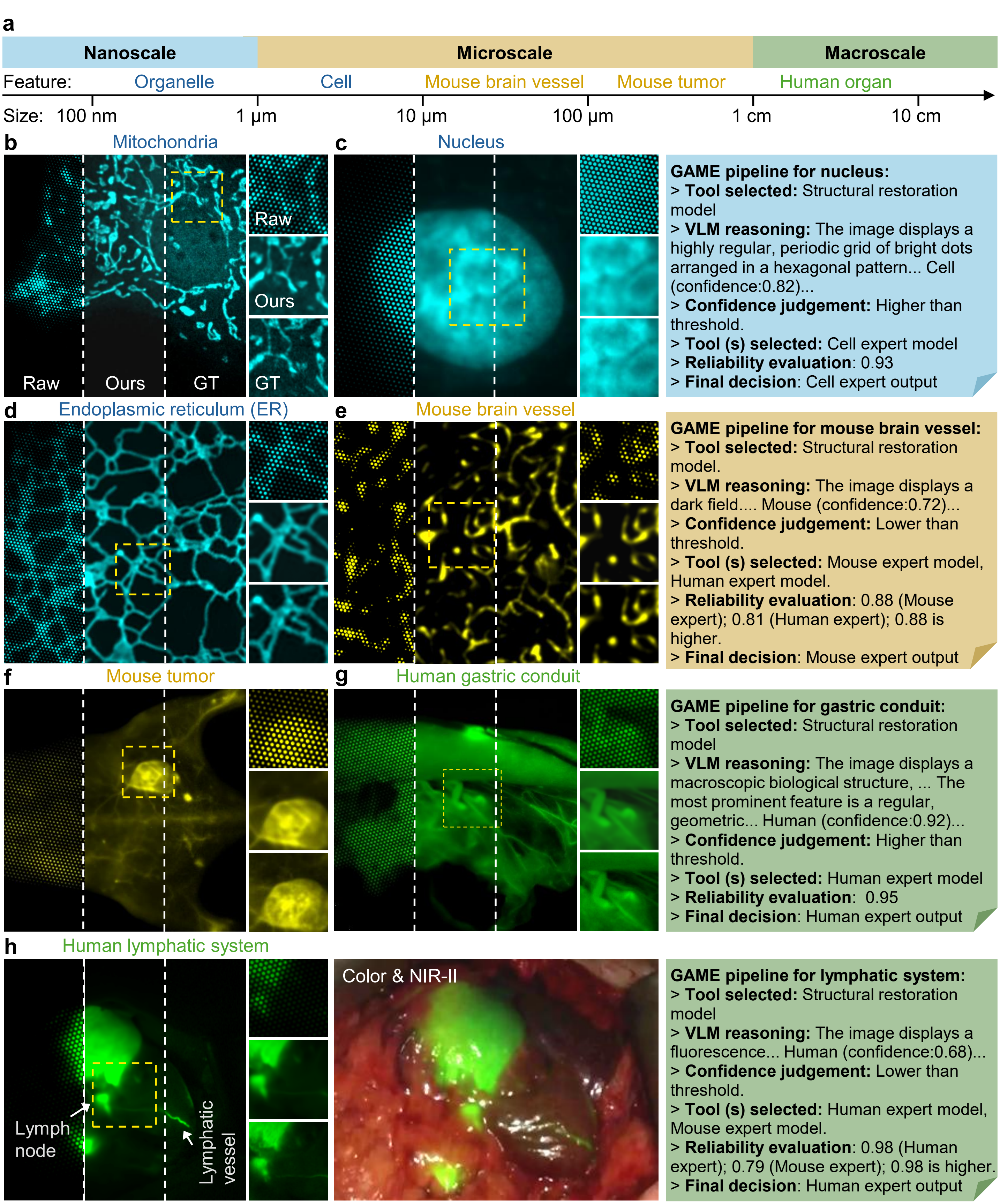
a
Nanoscale
Microscale
Macroscale
Feature:
Organelle
Cell
Mouse brain vessel
Mouse tumor
Human organ
Size: 100 nm
1 μm
10 μm
100 μm
1 cm
10 cm
b
Mitochondria
Raw
Ours
GT
c
Nucleus
GAME pipeline for nucleus:
> Tool selected: Structural restoration model
> VLM reasoning: The image displays a highly regular, periodic grid of bright dots arranged in a hexagonal pattern... Cell (confidence:0.82)...
> Confidence judgement: Higher than threshold.
> Tool (s) selected: Cell expert model
> Reliability evaluation: 0.93
> Final decision: Cell expert output
d
Endoplasmic reticulum (ER)
e
Mouse brain vessel
GAME pipeline for mouse brain vessel:
> Tool selected: Structural restoration model.
> VLM reasoning: The image displays a dark field.... Mouse (confidence:0.72)...
> Confidence judgement: Lower than threshold.
> Tool (s) selected: Mouse expert model, Human expert model.
> Reliability evaluation: 0.88 (Mouse expert); 0.81 (Human expert); 0.88 is higher.
> Final decision: Mouse expert output
f
Mouse tumor
g
Human gastric conduit
GAME pipeline for gastric conduit:
> Tool selected: Structural restoration model
> VLM reasoning: The image displays a macroscopic biological structure, ... The most prominent feature is a regular, geometric... Human (confidence:0.92)...
> Confidence judgement: Higher than threshold.
> Tool (s) selected: Human expert model
> Reliability evaluation: 0.95
> Final decision: Human expert output
h
Human lymphatic system
Lymph node
Lymphatic vessel
Color & NIR-II
GAME pipeline for lymphatic system:
> Tool selected: Structural restoration model
> VLM reasoning: The image displays a fluorescence... Human (confidence:0.68)...
> Confidence judgement: Lower than threshold.
> Tool (s) selected: Human expert model, Mouse expert model.
> Reliability evaluation: 0.98 (Human expert); 0.79 (Mouse expert); 0.98 is higher.
> Final decision: Human expert output

**Figure 5 | GAME-based cross-scale image restoration via autonomous expert routing.** (**a**) Overview of images containing objects with varying feature sizes used to evaluate the GAME network, spanning nanoscale organelles, microscale cellular and mouse-tissue structures, and macroscale mouse tissues and human organs. Representative restorations of subcellular structures, including (**b**) mitochondria, (**c**) nucleus, and (**d**) endoplasmic reticulum (ER). Restorations of mouse-tissue structures, such as (**e**) mouse brain vessels and (**f**) mouse tumors. Reconstruction of macroscale human anatomical structures, including (**g**) the human gastric conduit and (**h**) lymph nodes and lymphatic vessels. The restored NIR-II fluorescence endoscopic image of lymph nodes and lymphatic vessels is overlayed on the corresponding white-light intraoperative image for multimodal visualization. For each case, we compared the raw fiber-bundle images, GAME-restored outputs, ground-truth images, and their magnified ROIs. Dashed white lines separate image regions from different images, and yellow dashed boxes indicate magnified ROIs. The text boxes summarize the autonomous routing process of GAME, including VLM-based category recognition, confidence-guided expert selection, reliability evaluation, and final output determination. Higher reliability scores indicate better preservation of coarse structural information in the final expert reconstruction.

Supplementary Materials for

# Agentic AI-powered flexible fiber-bundle endoscopy for high-resolution NIR-II fluorescence imaging in vivo

Yanzhao Shi, Yuanhua Liu, Sixin Xu, Wayne Jason Li, Yuyuan Chen, Danyang Xu, Zhisheng Wu, Hanze Yu, Ian Yu-Hong Wong, Simon Ying-Kit Law, Hongjie Dai, Liangqiong Qu, and Feifei Wang

Correspondence to: hjdai@hku.hk, liangqqu@hku.hk, feifwang@hku.hk

## SUPPLEMENTARY FIGURES

**a**

| Fiber bundle | Core diameter (μm) | Pitch (μm) | NA | Wavelength (nm) | V | Mode number |
|---|---|---|---|---|---|---|
| **Fiber1** | 2.5 | 3.1 | 0.39 | 485 | 6.32 | ~20 |
| | | | | 850 | 3.60 | ~6 |
| | | | | 1050 | 2.92 | ~4 |
| | | | | 1350 | 2.27 | ~3 |
| | | | | 1550 | 1.98 | ~2 |
| **Fiber2** | 8.0 | 10.3 | 0.49 | 485 | 25.39 | ~322 |
| | | | | 850 | 14.49 | ~105 |
| | | | | 1050 | 11.73 | ~69 |
| | | | | 1350 | 9.12 | ~42 |
| | | | | 1550 | 7.95 | ~32 |
| **Fiber3** | 12.0 | 13.8 | 0.55 | 485 | 42.75 | ~914 |
| | | | | 850 | 24.39 | ~298 |
| | | | | 1050 | 19.75 | ~195 |
| | | | | 1350 | 15.36 | ~118 |
| | | | | 1550 | 13.38 | ~89 |

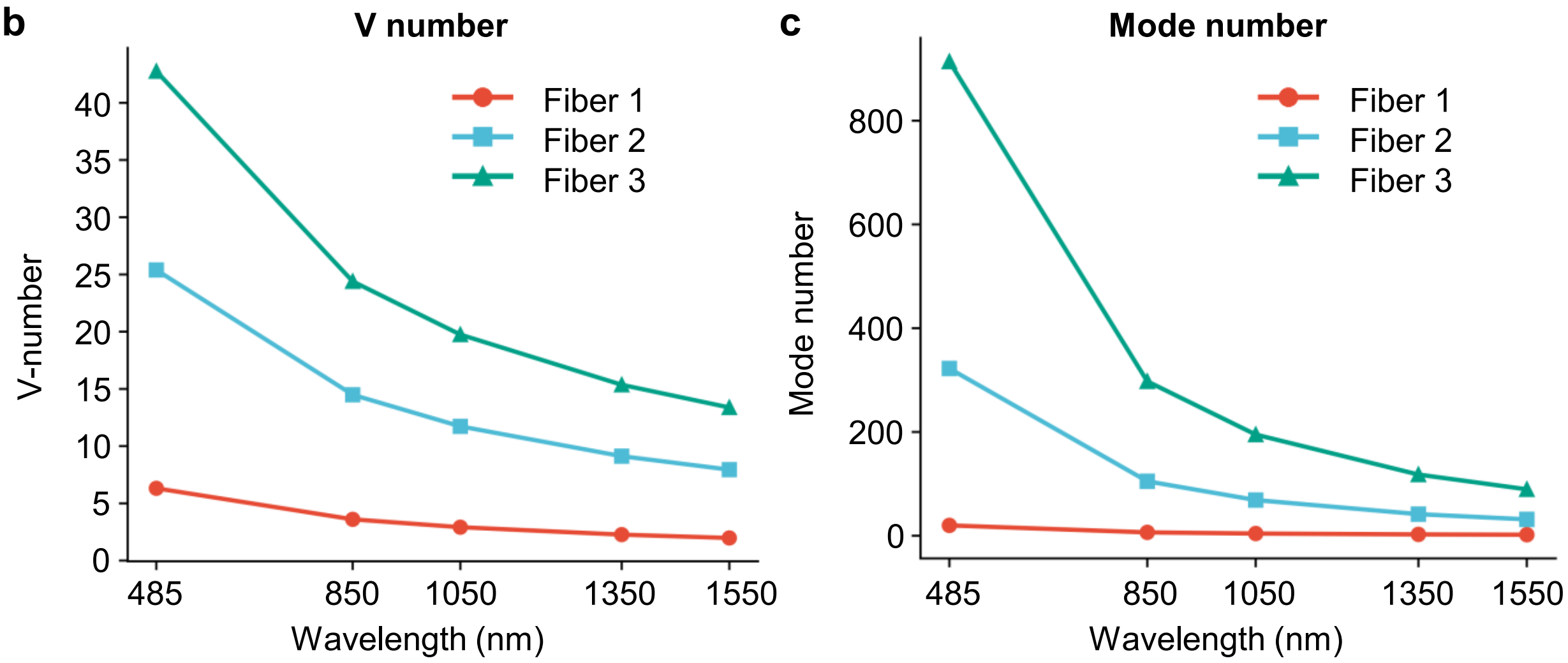

**Supplementary Figure 1. Fiber-bundle specifications and wavelength-dependent V-number and mode number.** (**a**) Optical and geometrical specifications of the three fiber bundles used in this study. Fiber 1 has a small core diameter and low numerical aperture (NA), whereas Fibers 2 and 3 were designed with larger core diameters, higher NAs, and increased core-to-core pitches to improve modal confinement and reduce wavelength-dependent inter-core crosstalk. The normalized frequency, or V-number, was calculated as $V = \pi D_{\mathrm{core}}\mathrm{NA}/\lambda$, where $D_{\mathrm{core}}$ is the fiber-

core diameter and $\lambda$ is the imaging wavelength. The approximate number of supported guided modes was estimated as $V^2/2$. Calculations were performed at 485, 850, 1050, 1350, and 1550 nm, corresponding to visible, NIR-I, and NIR-II imaging wavelengths. (**b**) Wavelength-dependent V-number for Fibers 1-3. The V-number decreases with increasing wavelength for all fiber bundles but remains substantially higher in Fibers 2 and 3 than in Fiber 1 owing to their larger core diameters and higher NAs. (**c**) Mode number as a function of wavelength for Fibers 1-3. Fiber 1 supports only a small number of guided modes at longer wavelengths, whereas Fibers 2 and 3 retain higher mode numbers across the visible-to-NIR-II spectral range. These calculations support the fiber-configuration strategy used to improve image transmission fidelity at longer wavelengths by maintaining stronger light confinement within individual fiber cores.

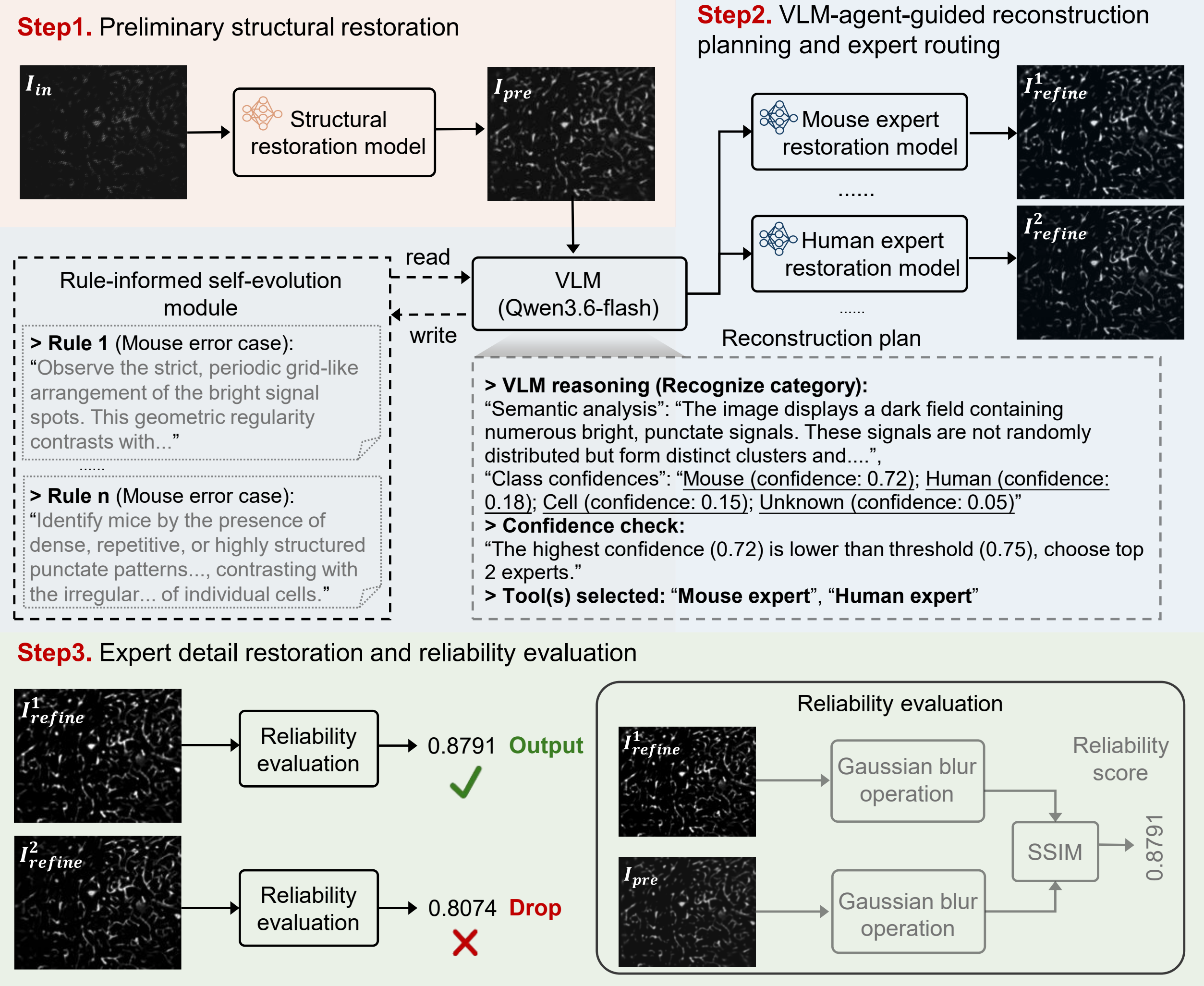

**Supplementary Figure 2. Detailed computational framework of the Agent-Guided Mixture-of-Experts (GAME) network.** Schematic of the GAME framework for fiber-bundle image restoration. The pipeline decomposes restoration into three sequential stages. In **Step 1**, a structural diffusion model performs preliminary structure restoration from the raw fiber-bundle image ($I_{\text{in}}$), removing honeycomb artifacts and generating a preliminary restored image ($I_{\text{pre}}$) that preserves the global biological morphology. In **Step 2**, a vision-language model (VLM) analyzes $I_{\text{pre}}$, assigns category confidence scores, and selects either one high-confidence expert or the top two experts when the prediction is uncertain. A rule-informed self-evolution module stores corrective routing rules to improve subsequent decisions. In **Step 3**, the selected expert diffusion models generate refined outputs, which are evaluated by comparing their low-frequency structural consistency with $I_{\text{pre}}$ after Gaussian blurring. The candidate with the higher SSIM-based reliability score is selected as the final output.

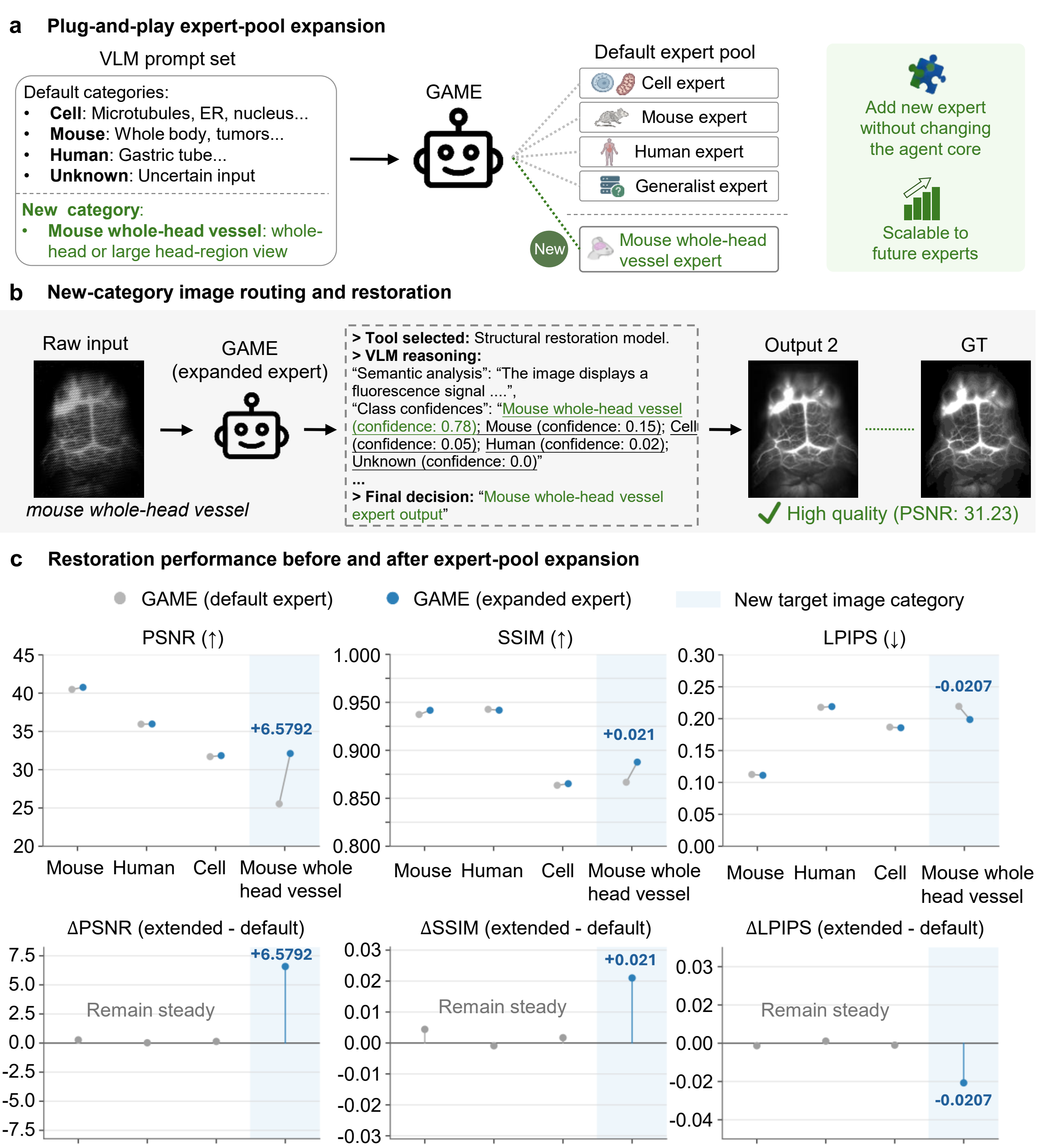

**Supplementary Figure 3. Plug-and-play expert-pool expansion of GAME for new-category image restoration.** (**a**) Schematic of the plug-and-play expert-pool expansion strategy. The default VLM prompt set contains Cell, Mouse (whole-body imaging), Human and Unknown categories, which correspond to the default expert pool. A new Mouse whole-head vessel category is introduced by updating the VLM prompt set and registering a newly trained Mouse whole-head vessel expert, without modifying the agent core. (**b**) Routing and restoration of a new-category image using the expanded expert pool. For an input image of mouse whole-head vessels, the VLM agent assigns the highest confidence score to the newly added Mouse whole-head vessel category and selects the corresponding expert for restoration. The restored output shows high agreement with the ground truth, with a PSNR of 31.23. (**c**) Quantitative comparison of restoration performance

before and after expert-pool expansion. Gray dots indicate GAME with the default expert pool, and blue dots indicate GAME with the expanded expert pool. The light-blue shaded region denotes the newly added target category. After expansion, restoration performance for the new Mouse whole-head vessel category improved substantially, with increases of 6.5792 in PSNR and 0.021 in SSIM and a decrease of 0.0207 in LPIPS, while performance on the original Mouse, Human and Cell categories remained stable. Higher PSNR and SSIM values and lower LPIPS values indicate better restoration quality.

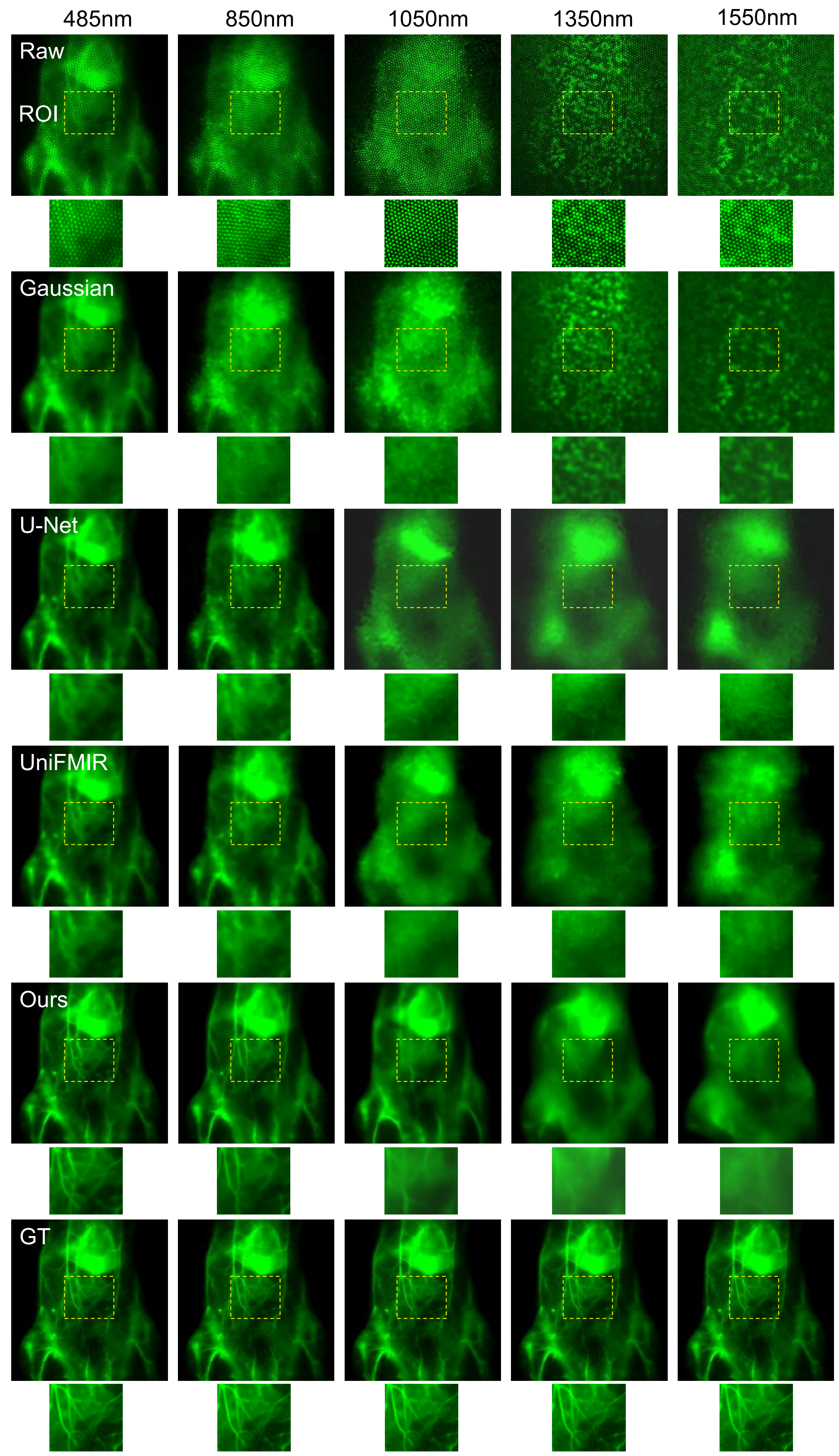

**Supplementary Figure 4. Comparison of different methods for restoring images transmitted through Fiber 1 at working wavelengths across the visible to NIR-II windows.** Rows show the

raw images transmitted through Fiber 1, Gaussian-filtered results, U-Net reconstructions, UniFMIR reconstructions, GAME reconstructions, and the corresponding ground-truth (GT) images. The GT image was a 1500-1700-nm NIR-IIb image showing blood vessels in a mouse. The raw images transmitted through Fiber 1 were generated using the DMD-based imaging system (Fig. 2b). Yellow dashed boxes mark regions of interest (ROIs), with enlarged views shown below each image. Fiber 1 showed progressively stronger image degradation at longer wavelengths due to wavelength-dependent inter-core crosstalk. GAME most effectively suppressed honeycomb artifacts and preserved vascular structures at 485 and 850 nm, but reconstruction quality decreased in the NIR-II range, where severe crosstalk limited the recoverable image information.

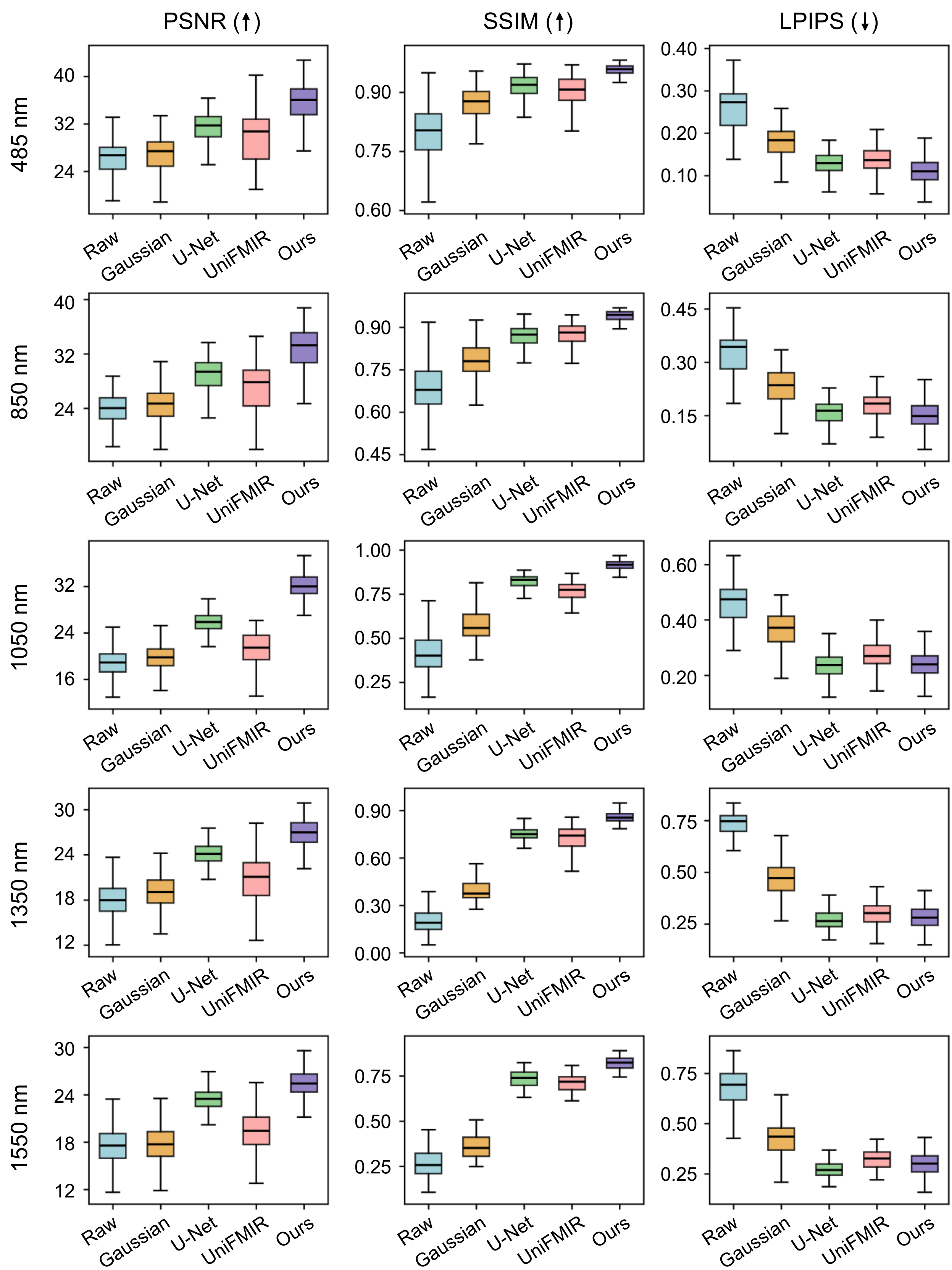

**Supplementary Figure 5. Quantitative comparison of different methods for restoring images transmitted through Fiber 1 at working wavelengths across the visible to NIR-II windows.** Restoration performance was compared among the raw input, Gaussian filtering, U-Net, UniFMIR,

and GAME using PSNR, SSIM, and LPIPS relative to the corresponding ground-truth images. Higher PSNR and SSIM values indicate better reconstruction fidelity, whereas lower LPIPS values indicate better perceptual similarity. GAME achieved consistently strong performance at 485-1050 nm, with improved artifact suppression and structural recovery compared with the baseline methods. At longer NIR-II wavelengths, severe crosstalk in Fiber 1 reduced the recoverable image information and narrowed the performance differences among the learning-based methods. Box plots summarize the metric distributions across the test images.

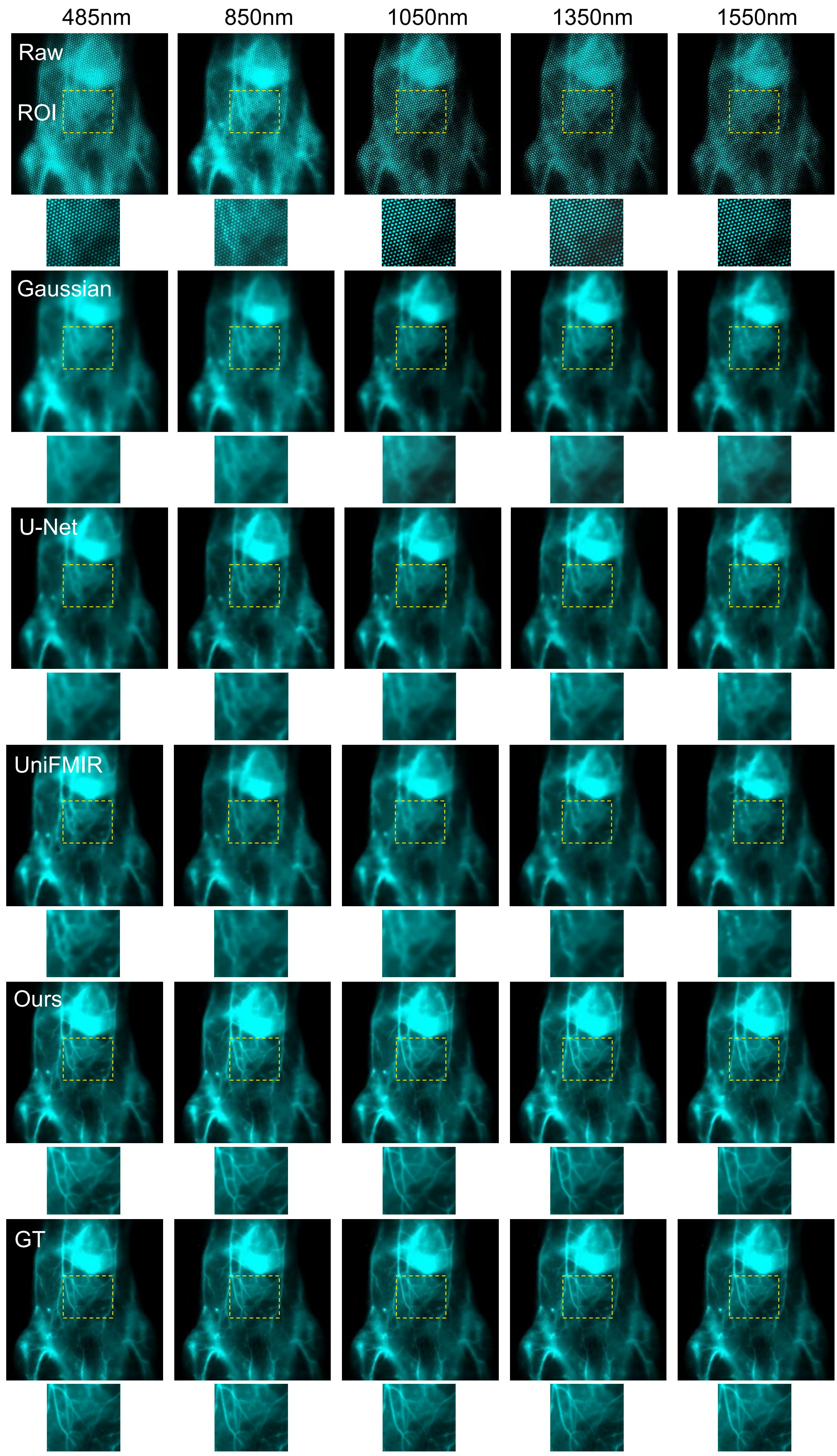

**Supplementary Figure 6. Comparison of different methods for restoring images transmitted through Fiber 2 at working wavelengths across the visible to NIR-II windows.** Rows show the

raw images transmitted through Fiber 2, Gaussian-filtered results, U-Net reconstructions, UniFMIR reconstructions, GAME reconstructions, and the corresponding GT images. The GT image is the same as that used in Supplementary Fig. 4. Yellow dashed boxes indicate ROIs, with enlarged views shown below each image. Compared with Fiber 1, the raw images transmitted through Fiber 2 preserved clearer vascular structures across the visible-to-NIR-II range, reflecting reduced wavelength-dependent inter-core crosstalk. GAME further suppressed honeycomb artifacts and restored fine vascular morphology with higher visual fidelity than baseline methods across all tested wavelengths.

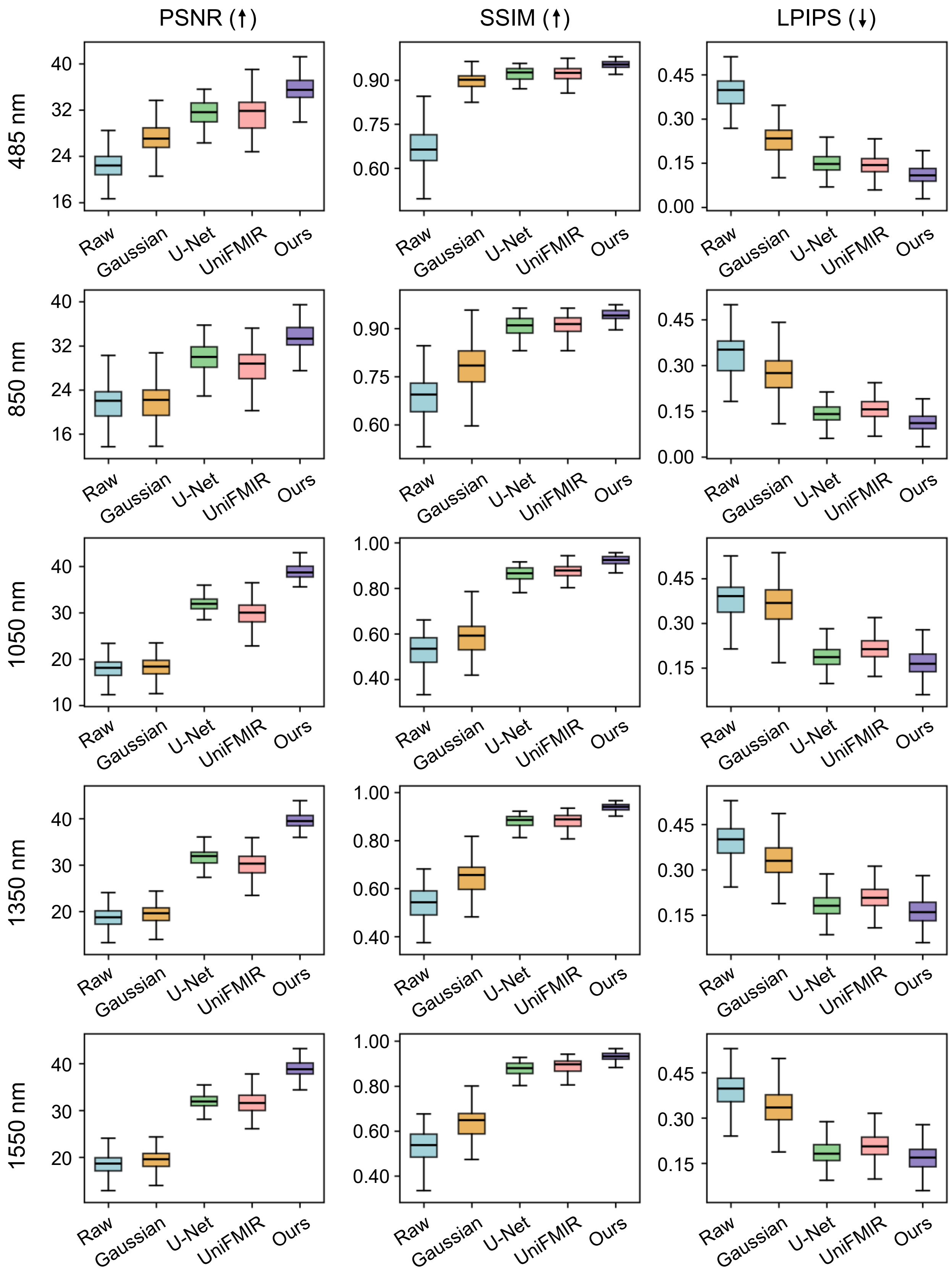

**Supplementary Figure 7. Quantitative comparison of different methods for restoring images transmitted through Fiber 2 at working wavelengths across the visible to NIR-II windows.** The raw input, Gaussian filtering, U-Net, UniFMIR, and GAME were compared using PSNR,

SSIM, and LPIPS relative to the corresponding ground-truth images. Higher PSNR and SSIM values indicate better reconstruction accuracy, whereas lower LPIPS values indicate higher perceptual similarity. GAME achieved strong and consistent performance across the visible-to-NIR-II range, demonstrating effective restoration under reduced wavelength-dependent crosstalk in Fiber 2. Box plots show the metric distributions across the test images.

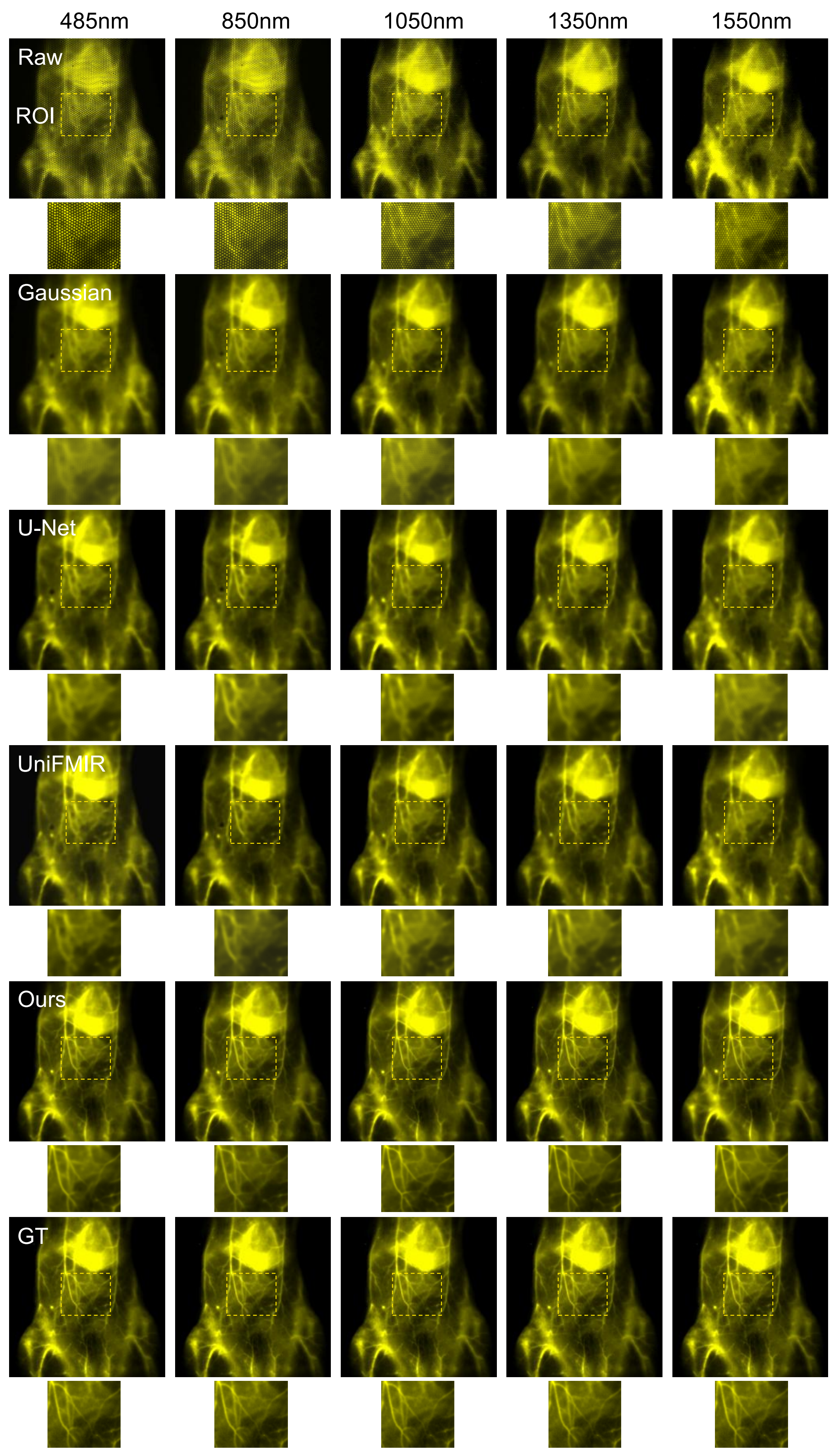

**Supplementary Figure 8. Comparison of different methods for restoring images transmitted through Fiber 3 at working wavelengths across the visible to NIR-II windows.** Rows show the

raw images transmitted through Fiber 3, Gaussian-filtered results, U-Net reconstructions, UniFMIR reconstructions, GAME reconstructions, and the corresponding GT images. The GT image is the same as that used in Supplementary Figs. 4 and 6. Yellow dashed boxes indicate ROIs, with enlarged views shown below each image. Owing to its larger core diameter, higher NA, and greater number of cores, Fiber 3 preserved vascular structures more effectively across the visible-to-NIR-II range. GAME further removed honeycomb artifacts and restored fine vascular features with high visual fidelity across all tested wavelengths.

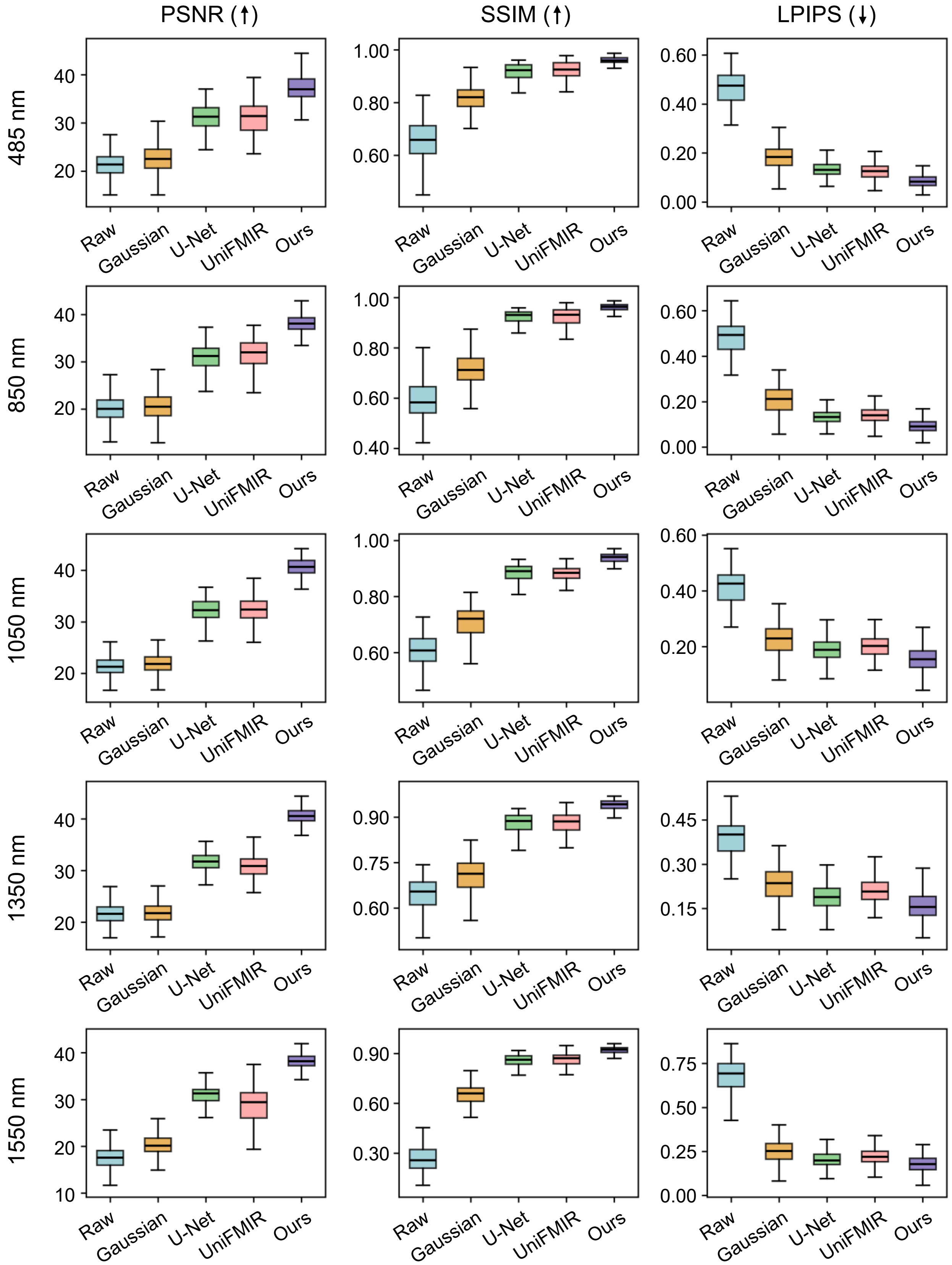

**Supplementary Figure 9. Quantitative comparison of different methods for restoring images transmitted through Fiber 3 at working wavelengths across the visible to NIR-II windows.** The raw input, Gaussian filtering, U-Net, UniFMIR, and GAME were compared using PSNR,

SSIM, and LPIPS relative to the corresponding ground-truth images. Higher PSNR and SSIM values indicate better reconstruction accuracy, whereas lower LPIPS values indicate higher perceptual similarity. GAME achieved the strongest overall performance across the visible-to-NIR-II range, consistent with the improved image transmission fidelity of Fiber 3 and its reduced wavelength-dependent degradation. Box plots show the metric distributions across the test images.

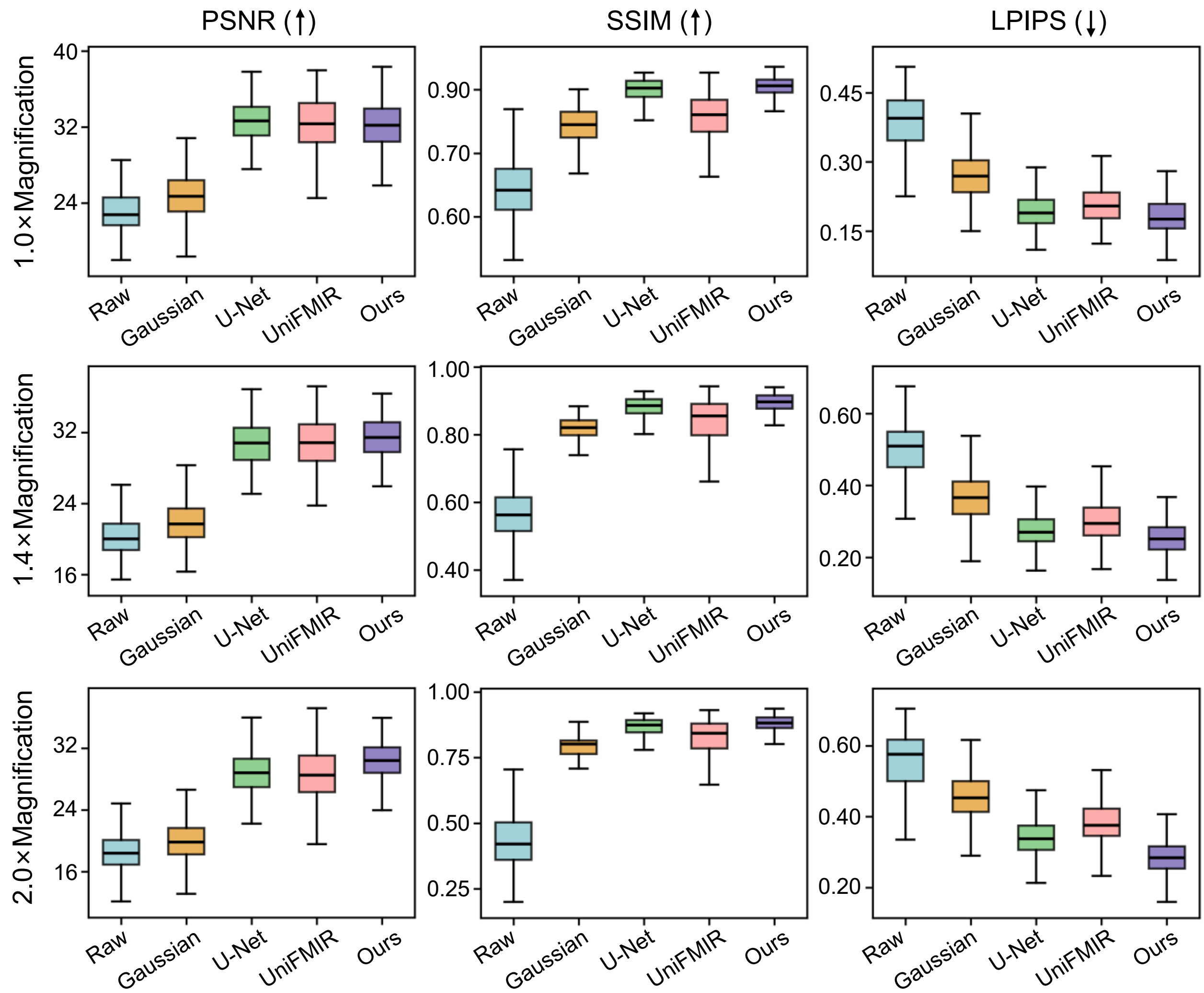

**Supplementary Figure 10. Quantitative evaluation of different methods for restoring images transmitted through Fiber 2 at 1550 nm at different image magnifications (Fig. 3).** Mouse vascular images at different magnifications were generated by scaling the ground-truth image input to the DMD by factors of 1.0×, 1.4×, and 2.0×. As the scaling factor increased, more fiber cores sampled the same vascular structure. The raw input, Gaussian filtering, U-Net, UniFMIR, and GAME were compared using PSNR, SSIM, and LPIPS relative to the corresponding ground-truth images. Higher PSNR and SSIM values indicate better reconstruction fidelity, whereas lower LPIPS values indicate higher perceptual similarity. GAME maintained strong restoration performance across all magnifications, demonstrating robust recovery of structural features under different effective sampling densities. Box plots show the metric distributions across the test images.

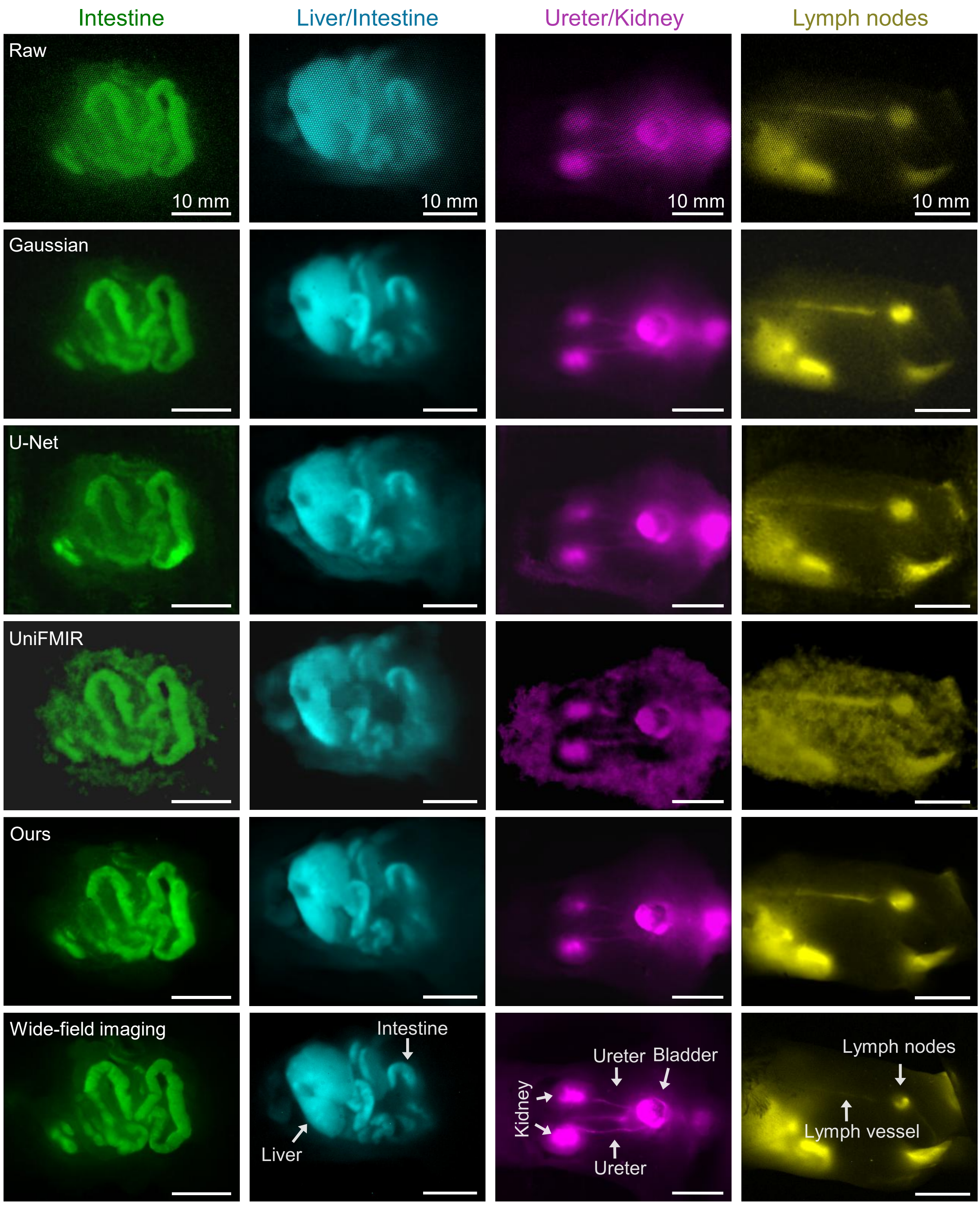

**Supplementary Figure 11. Comparison of restoration methods on unseen in vivo fiber-bundle endoscopic fluorescence images (Fig. 4).** Representative raw FBE images of ICG-labeled mouse intestine, liver, ureter, and lymphatic system were restored using Gaussian filtering, U-Net, UniFMIR, and our method. Wide-field fluorescence images of the corresponding anatomical regions are shown as clean references. Gaussian filtering and U-Net reduced honeycomb artifacts but introduced blurring or residual patterns, while UniFMIR showed less stable restoration on these out-of-distribution in vivo images. In contrast, our method more effectively suppressed fiber-bundle

artifacts while preserving organ boundaries and fine tubular structures, including the ureters and lymphatic vessels.

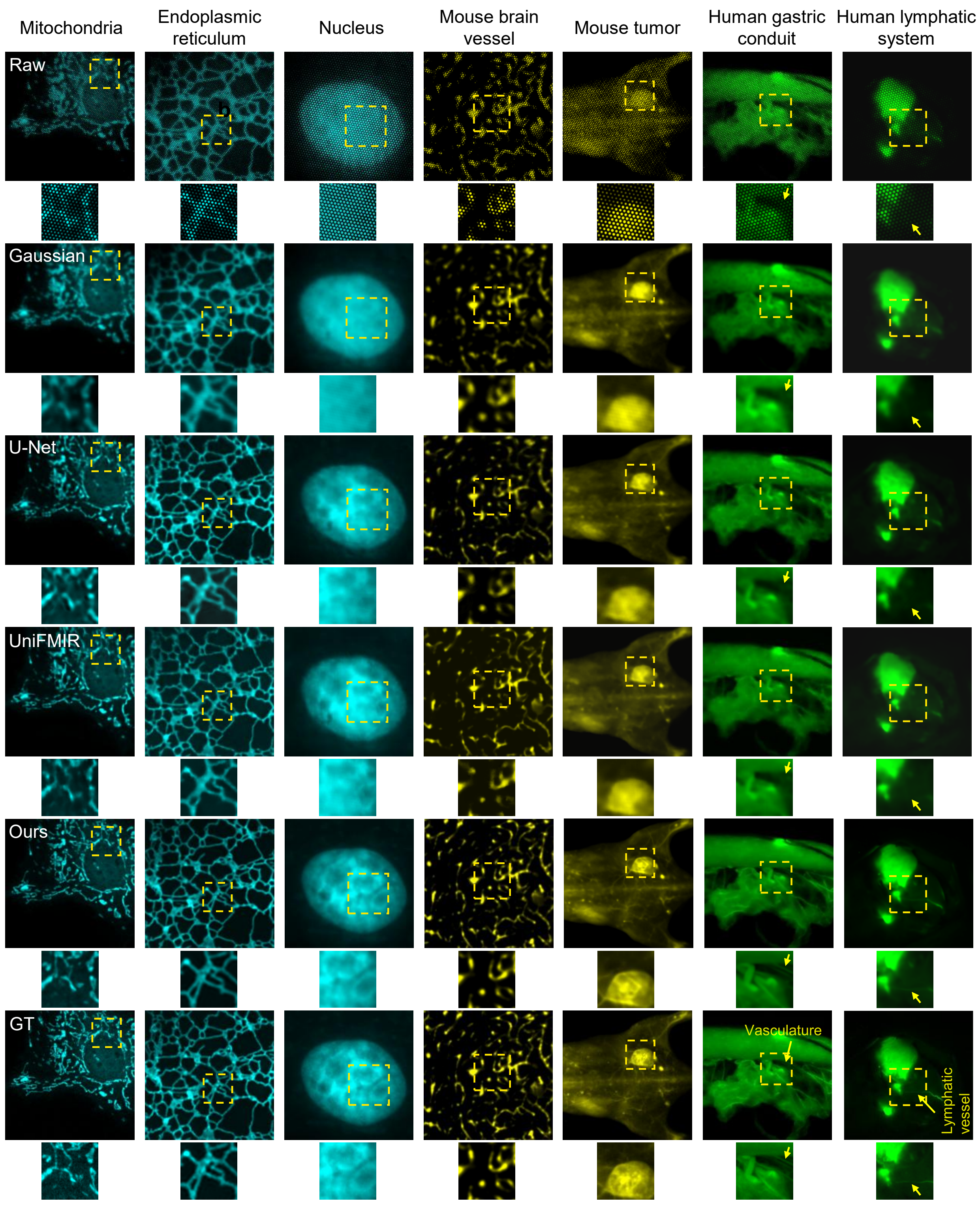

**Supplementary Figure 12. Comparison of different methods for restoring images containing biological objects with varying feature sizes (Fig. 5).** Columns show representative images of mitochondria, endoplasmic reticulum, nucleus, mouse brain vessels, mouse tumor, human lymphatic system, and human gastric conduit. Rows show the raw images transmitted through Fiber 2 at ~1550 nm, Gaussian-filtered results, U-Net reconstructions, UniFMIR reconstructions, GAME reconstructions, and the corresponding ground-truth or reference images. Yellow dashed boxes indicate ROIs, with enlarged views shown below each image. Compared with Gaussian filtering

and U-Net, GAME more effectively suppresses honeycomb artifacts while preserving fine structural details and global morphology across diverse biological targets.

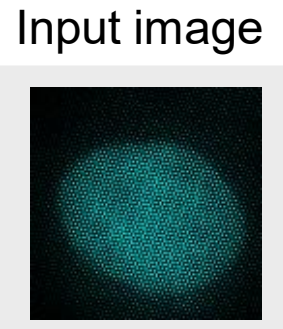
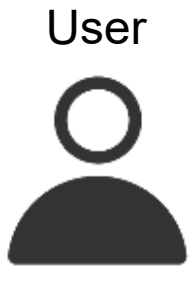
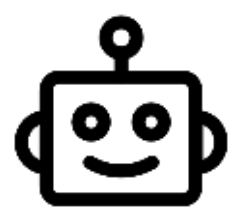
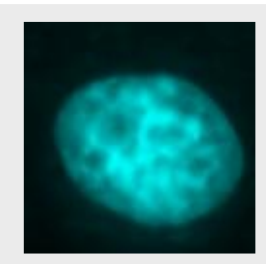
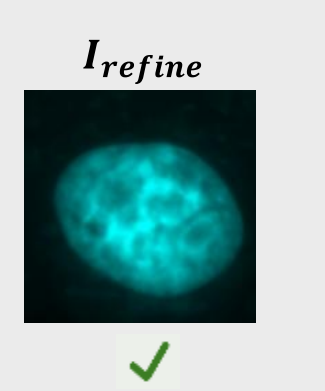

**Supplementary Figure 13. Detailed GAME routing example for images with high confidence scores.** Routing process for nucleus image restoration (Fig. 5c) using the GAME pipeline. In **Step 1**, the raw fiber-bundle input image is first processed by the structural restoration model to generate a preliminary restored image ($I_{\text{pre}}$). In **Step 2**, the VLM agent analyzes $I_{\text{pre}}$, recognizes a cellular pattern, and assigns the highest confidence score to the Cell category. In **Step 3**, because the Cell confidence score exceeds the predefined threshold of 0.75, GAME activates only the Cell expert for detail restoration. The refined output ($I_{\text{refine}}$) is further evaluated by comparing its Gaussian-blurred low-frequency structure with that of the preliminary restored image, $I_{\text{pre}}$, yielding a high SSIM-based reliability score of 0.9343. The Cell expert output is therefore accepted as the final reconstruction.

**Mouse brain vessel restoration**

Input image

User

GAME

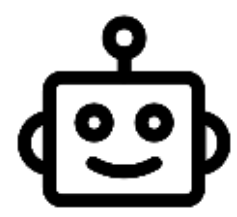

**Step1.** Preliminary structure restoration

> **Tool selected:** Structural restoration model
> **Preliminary structure restoring**...
> **Output** image $I_{pre}$ for step 2.

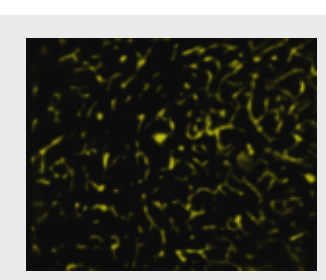

**Step2.** VLM-agent-guided reconstruction planning and expert routing

> **VLM reasoning:** ***Semantic analysis***: The image displays a dark field containing numerous bright, punctate signals. These signals are not randomly distributed but form distinct clusters and linear arrangements that resemble biological tissue architecture or vascular networks ... The density of the signals and their grouping into larger structural patterns distinguishes them from isolated individual cells ('Cell' category) or specific organ micrographs. The pattern is consistent with fluorescence imaging of a small animal subject (like a mouse) showing blood vessels or lymphatic structures. ***Class confidences***: Mouse (confidence:0.72), Human (confidence:0.18), Cell (confidence:0.15), Unknown (confidence:0.05).
> **Confidence check:** The highest confidence (0.72) is lower than threshold (0.75), choose top 2 experts.
> **Tool(s) selected:** Mouse expert, Human expert

**Step3.** Expert detail restoration and reliability evaluation

> **Mouse Expert restoring**...
> **Output** image $I^1_{refine}$
> **Human Expert restoring**...
> **Output** image $I^2_{refine}$
> **Reliability evaluation:**
Add gaussian blur to $I_{pre}$ and $I^1_{refine}$;
The SSIM of two blurred images is 0.8791.
The reliability score of mouse expert output is 0.8791.
Add gaussian blur to $I_{pre}$ and $I^2_{refine}$;
The SSIM of two blurred images is 0.8074.
The reliability score of human expert output is 0.8074.
Mouse expert's reliability is higher (0.8791 > 0.8074).
> **Final decision**: Mouse expert output

$I^1_{refine}$ ✓ $I^2_{refine}$ ✗

**Supplementary Figure 14. Detailed GAME routing example for an ambiguous image with a low confidence score.** GAME routing process for a morphologically ambiguous high-resolution mouse brain blood vessel image (Fig. 5e). In **Step 1**, the raw fiber-bundle input image is first processed by the structural restoration model to generate a preliminary restored image ($I_{\text{pre}}$). In **Step 2**, the VLM agent identifies the vascular structures but assigns a confidence score of 0.72 to the Mouse category, below the predefined threshold of 0.75, indicating uncertain classification. GAME therefore activates the top two candidate experts, the Mouse and Human experts, for parallel detail restoration. In **Step 3**, the refined outputs are evaluated by comparing their Gaussian-blurred low-frequency structures with that of $I_{\text{pre}}$ using SSIM. The Mouse expert output achieves a higher reliability score than the Human expert output (0.8791 versus 0.8074), and is selected as the final reconstruction.

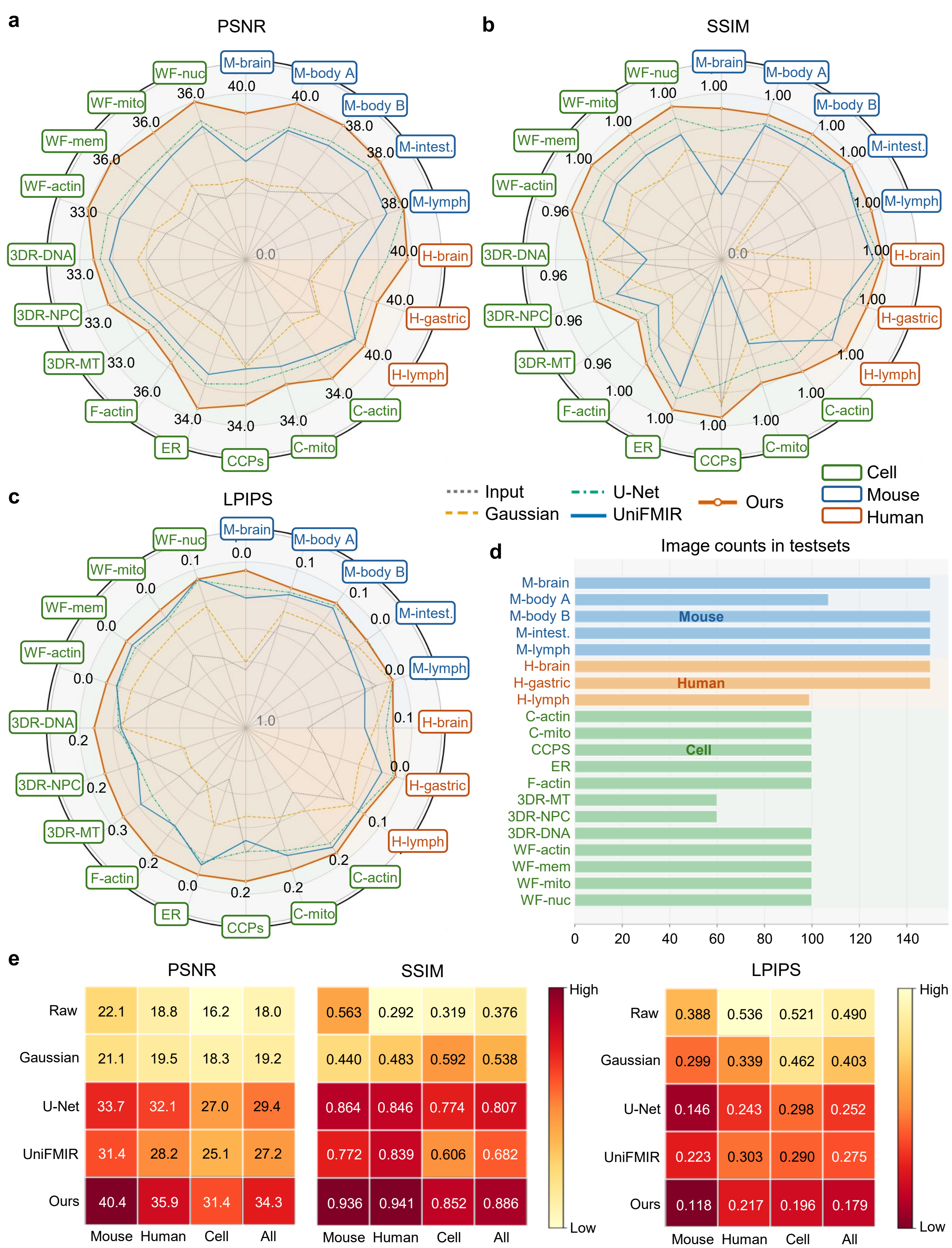

| PSNR | Mouse | Human | Cell | All |
|---|---|---|---|---|
| Raw | 22.1 | 18.8 | 16.2 | 18.0 |
| Gaussian | 21.1 | 19.5 | 18.3 | 19.2 |
| U-Net | 33.7 | 32.1 | 27.0 | 29.4 |
| UniFMIR | 31.4 | 28.2 | 25.1 | 27.2 |
| Ours | 40.4 | 35.9 | 31.4 | 34.3 |

| SSIM | Mouse | Human | Cell | All |
|---|---|---|---|---|
| Raw | 0.563 | 0.292 | 0.319 | 0.376 |
| Gaussian | 0.440 | 0.483 | 0.592 | 0.538 |
| U-Net | 0.864 | 0.846 | 0.774 | 0.807 |
| UniFMIR | 0.772 | 0.839 | 0.606 | 0.682 |
| Ours | 0.936 | 0.941 | 0.852 | 0.886 |

| LPIPS | Mouse | Human | Cell | All |
|---|---|---|---|---|
| Raw | 0.388 | 0.536 | 0.521 | 0.490 |
| Gaussian | 0.299 | 0.339 | 0.462 | 0.403 |
| U-Net | 0.146 | 0.243 | 0.298 | 0.252 |
| UniFMIR | 0.223 | 0.303 | 0.290 | 0.275 |
| Ours | 0.118 | 0.217 | 0.196 | 0.179 |

**Supplementary Figure 15. Quantitative comparison of GAME and baseline methods on diverse biological datasets.** (**a-c**) Radar plots comparing restoration performance across 20 test datasets spanning cell, mouse tissue and human tissue images using PSNR (**a**), SSIM (**b**) and LPIPS (**c**). Raw input, Gaussian filtering, U-Net, UniFMIR and GAME are compared. Dataset labels are color-coded by biological category: Cell, green; Mouse, blue; Human, orange. Mouse labels correspond to the 5 fine sub-datasets in the Mouse tissue dataset in Supplementary Table 3: M-brain, mouse brain vessel LSM; M-body A and B, mouse whole-body datasets A and B; M-intest., mouse intestine; and M-lymph, mouse lymph node. Human labels correspond to the 3 fine sub-datasets in the Human tissue dataset: H-brain, human brain vessel; H-gastric, human gastric tube; and H-lymph, human lymph node. Cell labels indicate public cell datasets: BioSR, including CCPs, ER, and F-actin for clathrin-coated pits, endoplasmic reticulum, and F-actin; 3DRCAN, including 3DR-MT, 3DR-NPC, and 3DR-DNA for microtubules, nuclear pore complexes and SiR-DNA-labeled DNA/nuclei, respectively; and GigaDB, including confocal actin and mitochondria images, denoted as C-actin and C-mito, and selected widefield subsets, denoted as WF-actin, WF-mito, WF-mem and WF-nuc, corresponding to 20× noise-level-1 actin, 20× noise-level-1 mitochondria, membrane and nucleus images, respectively. (**d**) Number of images in each test set used for quantitative evaluation. To avoid overweighting large test sets, up to 150 paired images were evaluated for each test set; for test sets containing fewer than 150 images, all available images were used. (**e**) Heat maps summarizing category-level and overall mean PSNR, SSIM and LPIPS for each method across Mouse, Human, Cell and all test sets. GAME showed the best overall performance, with an average PSNR of 34.3, SSIM of 0.886 and LPIPS of 0.179 across all test sets, outperforming Gaussian filtering, U-Net and UniFMIR. The improvements are consistent across biological scales. Higher PSNR and SSIM values and lower LPIPS values indicate better restoration quality.

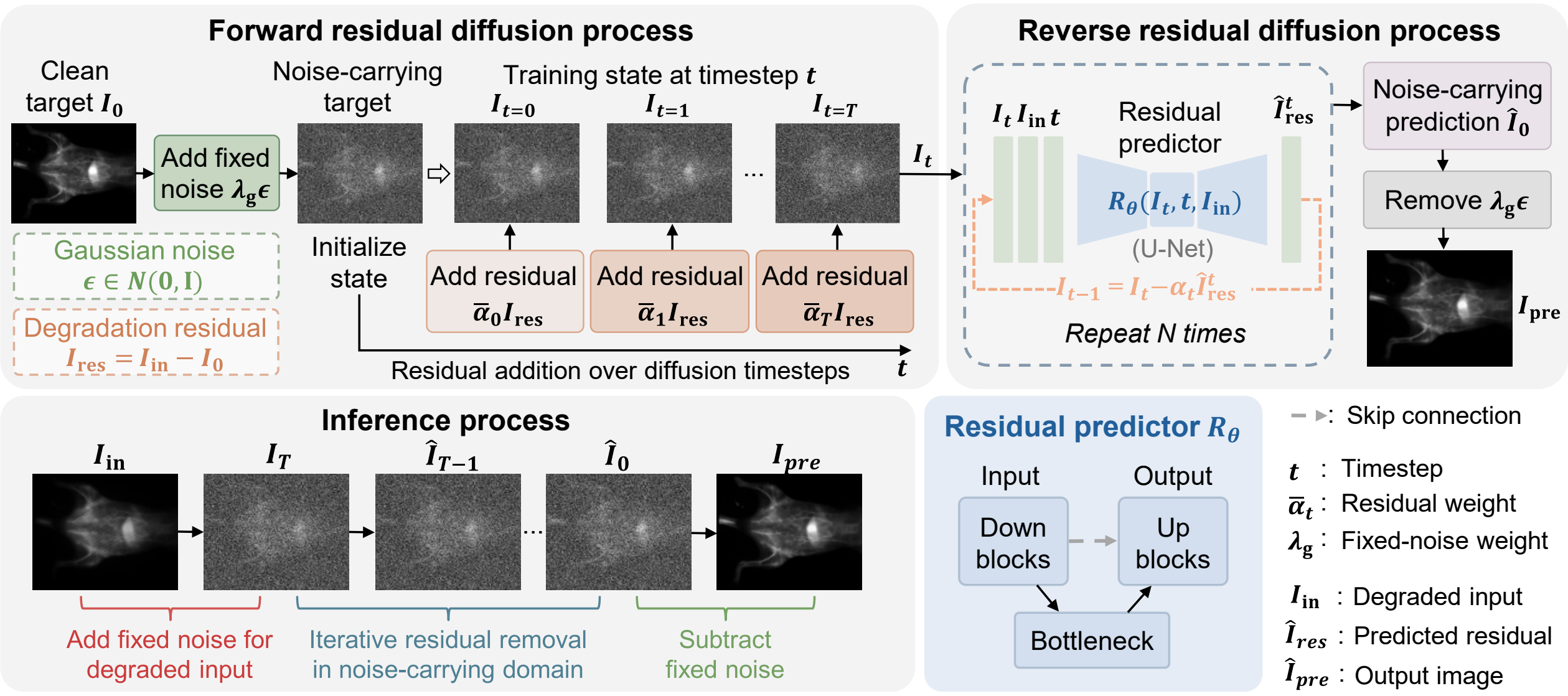

**Supplementary Figure 16. Detailed framework of the structural restoration model (SRM).** In the forward residual diffusion process, a clean target image $I_0$ is first lifted into a fixed noise-carrying domain by adding Gaussian noise, and the degradation residual $I_{res}$ is then progressively injected with timestep-dependent residual weights to generate training states $I_t$. In the reverse process, a time-conditioned residual predictor, implemented with a U-Net architecture, takes the current diffusion state $I_t$, the degraded input $I_{in}$, and the timestep $t$ as conditions to estimate the residual component. The predicted residual is iteratively removed in the noise-carrying domain, and the fixed Gaussian noise is subsequently subtracted to obtain the preliminary restored image. By learning the deterministic degradation residual rather than directly mapping the raw fiber-bundle image to the clean target, SRM suppresses dominant honeycomb artifacts and preserves reliable global morphology, providing a conservative structural prior for downstream VLM-guided expert routing and detail refinement in GAME.

**a** Automatic routing (Default mode)

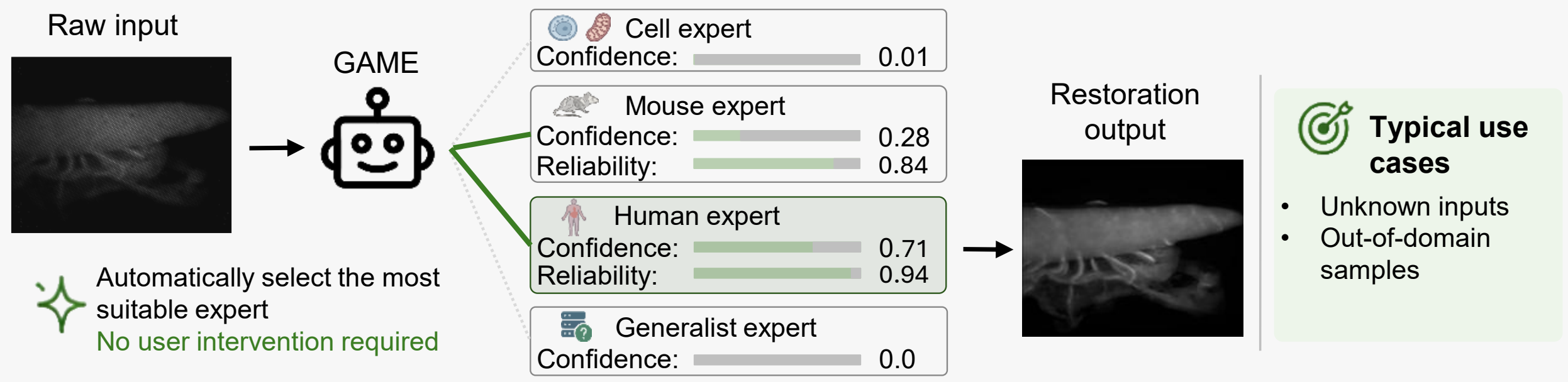

**b** User-guided routing (Use prior information)

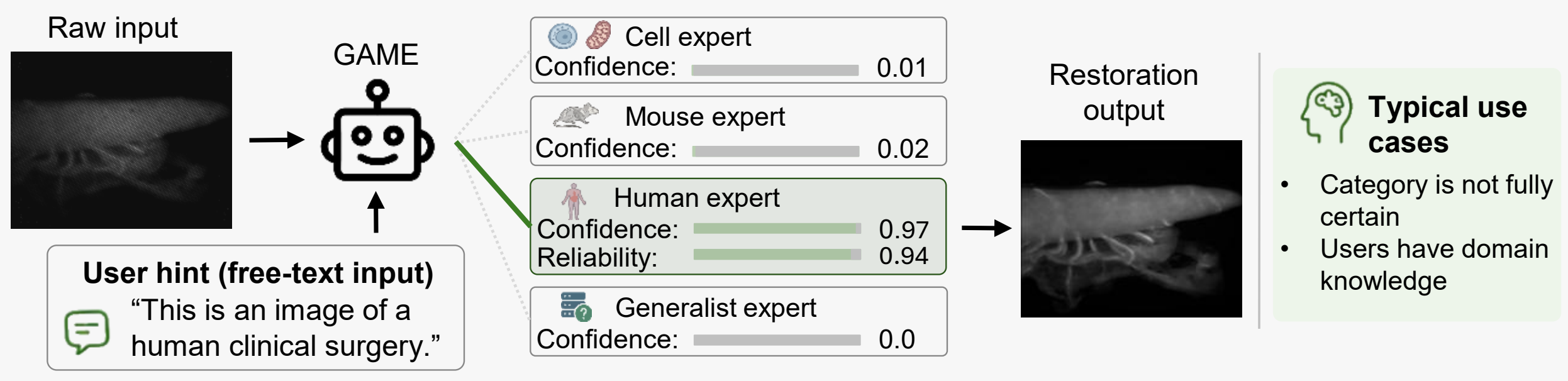

**c** Direct expert selection (Bypass routing)

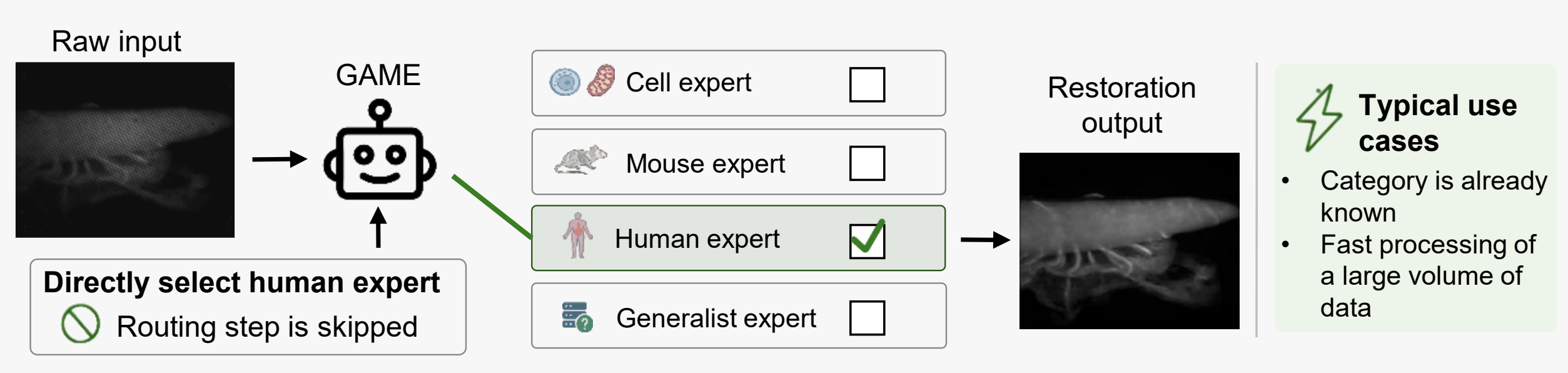

**Supplementary Figure 17. Flexible expert-selection modes in GAME pipeline.** (**a**) Automatic routing mode. The user provides only the raw fiber-bundle image, and GAME automatically evaluates candidate experts and selects the most suitable restoration pathway according to VLM-derived confidence scores and reliability evaluation. This default mode requires no user intervention and is suitable for unknown or out-of-domain inputs. (**b**) User-guided routing mode. The user provides a free-text hint as prior information, such as the clinical or biological context of the image. In the example shown, the hint indicating a human clinical surgery context increases the confidence assigned to the Human expert from 0.71 in (**a**) to 0.97 in (**b**), while decreasing the confidence of the Mouse expert from 0.28 to 0.02. This shift strengthens the separation between candidate experts and guides GAME toward a more informed expert-selection decision while using the same restoration pipeline. This mode is suitable when partial domain knowledge is available but the category is not fully certain. (**c**) Direct expert-selection mode. The user directly specifies the expert to be used, and VLM-based routing is bypassed. This mode is suitable when the image category is already known or when large batches of images from the same category require fast processing.

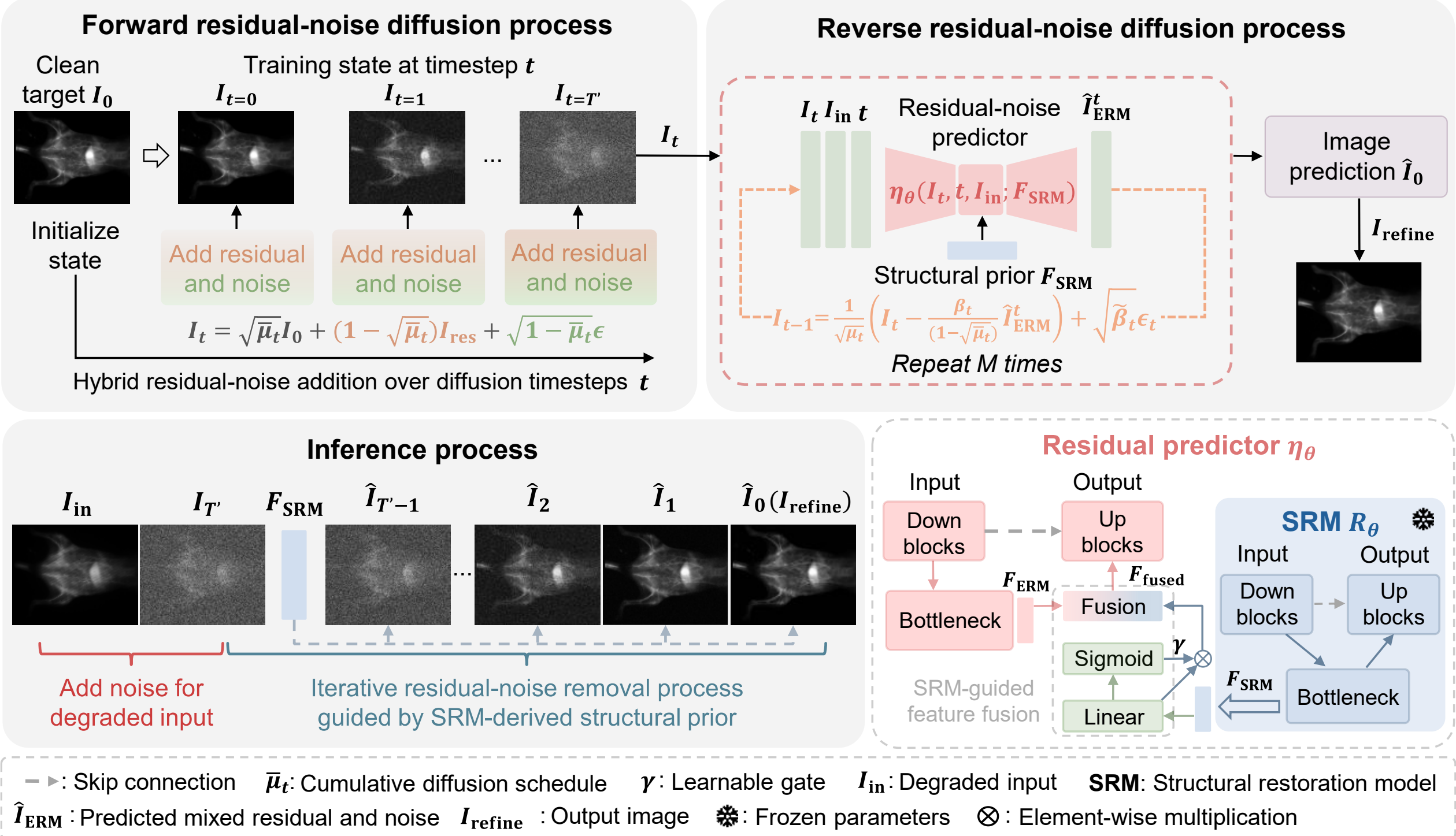

**Supplementary Figure 18. Detailed framework of the expert restoration model (ERM).** In the forward diffusion process, a clean target image $I_0$ is progressively transformed into training states $I_t$ by jointly injecting the degradation residual $I_{res} = I_{in} - I_0$ and Gaussian noise $\epsilon$ according to the cumulative diffusion schedule $\bar{\mu}_t$. In the corresponding reverse residual-noise diffusion process, inference starts from a noisy degraded input $I_{T'}$. At each reverse step, the residual-noise predictor uses the current diffusion state $I_t$, degraded input $I_{in}$, timestep $t$, and the SRM-derived structural prior to predict the residual-noise term $\hat{I}^t_{ERM}$, which is iteratively removed to generate the expert-refined output $I_{refine}$. The residual-noise predictor is implemented as a U-Net consisting of a downsampling block, a bottleneck, and an upsampling block, with an SRM-guided feature-fusion module. The ERM downsampling block, denoted as Down blocks in figure, first transforms the input diffusion state into restoration features $F_{ERM}$. Meanwhile, the frozen SRM branch provides intermediate structural features $F_{SRM}$ that encode the conservative global morphology. After being projected into the ERM feature space, this structural prior is adaptively fused with $F_{ERM}$ through a learnable gate $\gamma$ ,yielding the fused features. The fused features are then fed into the ERM upsampling blocks, denoted as Up blocks in figure, to predict the residual-noise term $\hat{I}^t_{ERM}$. This fusion strategy keeps the expert refinement anchored to reliable global morphology while allowing the ERM to restore expert-specific biological details, thereby reducing morphology-inconsistent hallucinations in the final output.

## SUPPLEMENTARY TABLES

### Supplementary Table 1. Diameter of commercial clinical endoscopes

| Brand | Category | Types included (Near-infrared fluorescence imaging) | Distal-end outer diameter range, mm (Distal-end outer diameter range, Fr) | Insertion-tube outer diameter range, mm (Insertion-tube outer diameter range, Fr) |
|---|---|---|---|---|
| Fujifilm | Gastroscope | Magnifying, therapeutic, ultrathin, and dual-channel gastroscopes (No) | 5.8-12.8 (17.4-38.4) | 5.9-12.8 (17.7-38.4) |
| | Double-balloon enteroscope | Standard, ultrathin, and short double-balloon enteroscopes (No) | 7.5-9.4 (22.5-28.2) | 7.7-9.3 (23.1-27.9) |
| | Duodenoscope | Duodenoscopes (No) | 13.1 (39.3) | 11.3 (33.9) |
| | Colonoscope | Standard, magnifying, ultrathin, therapeutic, Vision Assist, and G-EYE colonoscopes (No) | 9.8-12.8 (29.4-38.4) | 10.7-12.8 (32.1-38.4) |
| | Bronchoscope | Diagnostic, therapeutic, and advanced diagnostic bronchoscopes (No) | 3.8-5.8 (11.4-17.4) | 3.8-6.1 (11.4-18.3) |
| Olympus | Cystoscope | Diagnostic cystoscopes with HD imaging, suction, retroflexion, and wide-field configurations (No) | 2.7-4.6 (8.1-13.8) | 5.5 (16.5) |
| | Rhinoscope | Diagnostic rhinoscopes with wide-field, ultrathin, HD, portable, and high-resolution configurations (No) | 1.8-4.8 (5.4-14.4) | 2.2-5.0 (6.6-15.0) |
| | Gastroscope | Diagnostic gastroscopes with HD imaging and near-focus mode (No) | 9.9 (29.7) | 9.6 (28.8) |
| | Hysteroscope | Flexible ultrathin diagnostic hysteroscopes (No) | 3.0 (9.0) | 3.1 (9.3) |
| | Ureteroscope | Flexible fiber and video ureterorenoscopes (No) | 1.63-2.83 (4.89-8.49) | 2.65-3.30 (7.95-9.90) |
| Richard Wolf | Urethrocystoscope | Flexible video urethrocystoscopes (No) | 3.6 (10.8) | 5.3 (15.9) |

| **Brand** | **Category** | **Types included**<br>**(Near-infrared fluorescence imaging)** | **Distal-end outer diameter range, mm (Distal-end outer diameter range, Fr)** | **Insertion-tube outer diameter range, mm**<br>**(Insertion-tube outer diameter range, Fr)** |
|---|---|---|---|---|
| | Choledochoscope | Flexible fiber and sensor choledochoscopes<br>(No) | 5.0-5.4 (15.0-16.2) | - |
| | Hysteroscope | Flexible fiber hysteroscopes<br>(No) | 2.5 (7.5) | 3.6 (10.8) |
| Storz | Ureteroscope | Flexible Ureteroscope<br>(No) | 3.5-3.6 (10.5-10.8) | 7.5-9.0 (22.5-27.0) |
| | Hysteroscope | Hysteroscope systems<br>(Yes) | 2.0-3.6 (6.0-10.8) | 3.7-5.0 (11.1-15.0) |
| | Laparoscopy | Fluorescence laparoscopy systems<br>(Yes) | - | 5-10 (15-30) |
| | Urethrocystoscope | Flexible video urethrocystoscopes<br>(No) | 3.6 (10.8) | 5.3 (15.9) |
| Terumo | Falloposcope | Flexible fiber falloposcope<br>(No) | 0.5-0.6 (1.5-1.8) | - |
| FemDx Medsystems | Falloposcope | Disposable hand-held CMOS falloposcope<br>(No) | 1.2 (3.6) | - |

**Supplementary Table 2. Reported flexible endoscopic imaging modalities**

| Fiber imaging type | FOV | Resolution | Fluorescence imaging | Fluorescence wavelength (excitation/emission) | References |
|---|---|---|---|---|---|
| Scanning single-mode fiber endoscopy | / | Lateral: 20 µm | Yes | 633 nm / ~645 nm | Seibel & Smithwick (2002) [1] |
| | / | / | Yes | 405 nm / 420-600 nm; 405 nm / 620-650 nm; 405 nm / 650-700 nm | Seibel et al. (2008) [2] |
| | / | / | Yes | 442 nm / Not reported; 532 nm / Not reported | Lee et al. (2010) [3] |
| | / | Lateral: 10 µm | Yes | 815 nm / Not reported | Seibel et al. (2006) [4] |
| | 600 × 500 µm | Lateral: 1-5 µm; Axial: 5-15 µm | Yes | 488 nm / ~500-650 nm | Osdoit et al. (2007) [5] |
| | 390×390 µm | Pixel size: 0.8 µm | Yes | 488 nm / Not reported | Thong et al. (2007) [6] |
| | 110×110 µm | Lateral: 0.8 µm; Axial: 10 µm | Yes | 800 nm / Not reported | Rivera et al. (2011) [7] |
| | ~110 µm diameter | Lateral: ~0.76 µm; Axial: ~4.36 µm | No | / | Zhang et al. (2012) [8] |
| | / | Lateral: ~12 µm | Yes | 424 nm / ~456-484 nm; 488 nm / ~508-554 nm; 642 nm / >647 nm | Savastano et al. (2017) [9] |
| | 378×439 µm | / | No | / | Seo et al. (2018) [10] |
| | Lateral scanning range: 3 mm | Lateral: ~14-17 µm; Axial: ~9 µm | No | / | Park et al. (2019) [11] |
| | 350×350 µm | Lateral: 3.2 µm | Yes | 785 nm / Not reported | Hwang et al. (2020) [12] |
| Multimode fiber endoscopy | 80×80 µm | Lateral: 2 µm | Yes | / | Caravaca-Aguirre & Piestun (2017) [13] |
| | / | / | / | / | Rahmani et al. (2018) [14] |
| | ~145 µm lateral FOV | / | Yes | 488 nm / Not reported; 640 nm / Not reported | Wen et al. (2023) [15] |
| | 32 × 32 µm | Lateral: ~0.716 µm | Yes | 405 nm / 446-486 nm; 405 nm / 515-565 nm | Fay et al. (2025) [16] |

| Fiber imaging type | FOV | Resolution | Fluorescence imaging | Fluorescence wavelength (excitation/emission) | References |
|---|---|---|---|---|---|
| Flexible fiber-bundle endoscopy | / | Lateral: ~127 µm | / | / | Hopkins & Kapany (1954) [17] |
| | / | / | / | / | Hirschowitz et al. (1957) [18] |
| | / | / | / | / | Marshall (1964) [19] |
| | / | Lateral: 3.2 µm; axial: 20 µm | Yes | 880 nm / Not reported; | Göbel et al. (2004) [20] |
| | 350-750 µm diameter | Lateral: 4.4 µm | Yes | 455 nm / Not reported; | Muldoon et al. (2007) [21] |
| | 230 µm diameter | Lateral: 1.4 µm; axial: ~26 µm | Yes | 561 nm / ~631 nm (> 580 nm) | Lane et al. (2009) [22] |
| | 600 µm diameter | / | Yes | 445-485 nm / 532-600 nm | Orth et al. (2019) [23] |
| | / | / | Yes | / | Nagengast et al. (2019) [24] |
| | / | Lateral: 1-4.32 µm | No | / | Sun et al. (2022) [25] |
| | / | Lateral: 0.85 µm; axial: 14 µm | No | / | Choi et al. (2022) [26] |
| | ~250 µm | / | No | / | Badt & Katz (2022) [27] |
| | 333 µm diameter | Lateral: 3.91-6.20 µm; axial: 35.4 µm | Yes | 473 nm / 525 nm | Zhou et al. (2022) [28] |
| | / | / | Yes | ~750 nm / 780-900 nm | Stibbe et al. (2023) [29] |
| | 230 µm diameter | Lateral: ~2 µm | Yes | 488 nm / Not reported; | Du et al. (2024) [30] |
| | ~650 µm diameter | Lateral: 548 nm (best case) | No | / | Song et al. (2024) [31] |
| | ~2.5-9 mm diameter | Lateral : ≤100 µm; axial : 10.7µm | Yes | 405 nm / Not reported;<br>488 nm / Not reported;<br>520 nm / Not reported;<br>642 nm / Not reported | Kiekens et al. (2020) [32] |
| | / | Lateral: ~133-154 µm | Yes | ~785 nm / ~835-875 nm | Gallagher et al. (2025) [33] |

**Supplementary Table 3. Summary of paired fiber-bundle image datasets**

| Dataset block | Biological category | Fiber bundle | Wavelength (nm) | Fine sub-datasets | Train and validation | Test | Total pairs |
|---|---|---|---|---|---|---|---|
| Fiber 1 wavelength dataset (generated using the DMD system shown in Fig. 2b) | Mouse | Fiber 1 | 485 | 485-nm-data | 900 | 100 | 5,000 |
| | Mouse | Fiber 1 | 850 | 850-nm-data | 900 | 100 | |
| | Mouse | Fiber 1 | 1050 | 1050-nm-data | 900 | 100 | |
| | Mouse | Fiber 1 | 1350 | 1350-nm-data | 900 | 100 | |
| | Mouse | Fiber 1 | 1550 | 1550-nm-data | 900 | 100 | |
| Fiber 2 wavelength dataset (generated using the DMD system shown in Fig. 2b) | Mouse | Fiber 2 | 485 | 485-nm-data | 900 | 100 | 5,000 |
| | Mouse | Fiber 2 | 850 | 850-nm-data | 900 | 100 | |
| | Mouse | Fiber 2 | 1050 | 1050-nm-data | 900 | 100 | |
| | Mouse | Fiber 2 | 1350 | 1350-nm-data | 900 | 100 | |
| | Mouse | Fiber 2 | 1550 | 1550-nm-data | 900 | 100 | |
| Fiber 3 wavelength dataset (generated using the DMD system shown in Fig. 2b) | Mouse | Fiber 3 | 485 | 485-nm-data | 900 | 100 | 5,000 |
| | Mouse | Fiber 3 | 850 | 850-nm-data | 900 | 100 | |
| | Mouse | Fiber 3 | 1050 | 1050-nm-data | 900 | 100 | |
| | Mouse | Fiber 3 | 1350 | 1350-nm-data | 900 | 100 | |
| | Mouse | Fiber 3 | 1550 | 1550-nm-data | 900 | 100 | |
| Mouse FOV dataset (generated using the DMD system shown in Fig. 2b) | Mouse | Fiber 2 | 1550 | Magnification 1.0 | 1,620 | 182 | 10,812 |
| | Mouse | Fiber 2 | 1550 | Magnification 1.2 | 1,620 | 182 | |
| | Mouse | Fiber 2 | 1550 | Magnification 1.4 | 1,620 | 182 | |
| | Mouse | Fiber 2 | 1550 | Magnification 1.6 | 1,620 | 182 | |
| | Mouse | Fiber 2 | 1550 | Magnification 1.8 | 1,620 | 182 | |
| | Mouse | Fiber 2 | 1550 | Magnification 2.0 | 1,620 | 182 | |
| Mouse tissue dataset (generated using the DMD system shown in Fig. 2b) | Mouse | Fiber 2 | 1550 | Brain vessel LSM | 1,000 | 600 | 8,466 |
| | Mouse | Fiber 2 | 1550 | Whole body A | 1,840 | 214 | |
| | Mouse | Fiber 2 | 1550 | Whole body B | 1,620 | 182 | |
| | Mouse | Fiber 2 | 1550 | Intestine | 1,100 | 220 | |
| | Mouse | Fiber 2 | 1550 | Lymph node | 1,300 | 390 | |

| Dataset block | Biological category | Fiber bundle | Wavelength (nm) | Fine sub-datasets | Train and validation | Test | Total pairs |
|---|---|---|---|---|---|---|---|
| Human tissue dataset (generated using the DMD system shown in Fig. 2b) | Human | Fiber 2 | 1550 | Brain vessel | 1,000 | 165 | 3,657 |
| | Human | Fiber 2 | 1550 | Gastric tube | 1,688 | 170 | |
| | Human | Fiber 2 | 1550 | Lymph node | 535 | 99 | |
| Cell dataset (generated using the DMD system shown in Fig. 2b) | Cell | Fiber 2 | 1550 | 3DRCAN | 8,410 | 220 | 20,020 |
| | Cell | Fiber 2 | 1550 | BioSR | 2,490 | 400 | |
| | Cell | Fiber 2 | 1550 | GigaDB 100888, confocal | 1,220 | 200 | |
| | Cell | Fiber 2 | 1550 | GigaDB 100888, widefield | 6,280 | 800 | |
| Extended diverse dataset (generated using the DMD system shown in Fig. 2b) | Mouse, Human, Cell | Fiber 2 | 1550 | Extended-data | 3,429 | 1,144 | 4,573 |
| Endoscopy dataset (screen-displayed images captured through a lens-coupled fiber bundle endoscope) | Mouse | Fiber 2 | / | Whole body A | 920 | 107 | 5,587 |
| | Mouse | Fiber 2 | / | Whole body B | 900 | 100 | |
| | Mouse | Fiber 2 | / | Intestine | 1,100 | 220 | |
| | Mouse | Fiber 2 | / | Lymph node | 1,300 | 390 | |
| | Mouse | Fiber 2 | / | Ureter | 500 | 50 | |
| Endoscopy wavelength dataset (DMD-displayed images captured through a lens-coupled fiber bundle endoscope) | Mouse | Fiber 2 | 485 | 485-nm-data | 4,720 | 867 | 27,935 |
| | Mouse | Fiber 2 | 850 | 850-nm-data | 4,720 | 867 | |
| | Mouse | Fiber 2 | 1050 | 1050-nm-data | 4,720 | 867 | |
| | Mouse | Fiber 2 | 1350 | 1350-nm-data | 4,720 | 867 | |
| | Mouse | Fiber 2 | 1550 | 1550-nm-data | 4,720 | 867 | |
| Total | / | / | / | / | 83,452 | 12,598 | 96,050 |

**Supplementary References:**